\documentclass[twoside,11pt]{article}

\usepackage{jmlr2e}
\DeclareMathAlphabet{\pazocal}{OMS}{zplm}{m}{n}

\usepackage{stmaryrd}
\usepackage{bbm}
\usepackage{multirow}
\usepackage{pifont}
\usepackage{textcomp}
\usepackage{dsfont}
\usepackage{amsmath}
\usepackage{amssymb}
\usepackage{graphicx}   
\usepackage{verbatim}   
\usepackage{color}      
\usepackage{subfig}  
\usepackage{algorithm}
\usepackage{algorithmic}
\usepackage{hyperref}   

\def\NN{{\mathbb N}}
\def\PP{{\mathbb P}}

\def\Acal{{\mathcal A}}
\def\Pcal{{\mathcal P}}

\def\Ocal{{\mathcal O}}

\def\Scal{{\mathcal S}}

\def\Mcal{{\mathcal M}}

\def\Xcal{{\mathcal X}}

\def\Vcal{{\mathcal V}}

\def\kl{\texttt{KL}}
\def\klber{\texttt{kl}}

\def\bsym{\boldsymbol}

\newcommand{\sk}{\nonumber\\}

\newcommand{\bp}{\noindent{\emph{Proof}.}\ }
\newcommand{\ep}{\hfill $\Box$}
\newcommand{\BEAS}{\begin{eqnarray*}}
\newcommand{\EEAS}{\end{eqnarray*}}
\newcommand{\BEA}{\begin{eqnarray}}
\newcommand{\EEA}{\end{eqnarray}}
\newcommand{\BEQ}{\begin{equation}}
\newcommand{\EEQ}{\end{equation}}
\newcommand{\BIT}{\begin{itemize}}
\newcommand{\EIT}{\end{itemize}}
\newcommand{\BNUM}{\begin{enumerate}}
\newcommand{\ENUM}{\end{enumerate}}

\newcommand{\argmax}{\arg\!\max}
\newcommand{\beq}{\begin{equation}}
\newcommand{\eeq}{\end{equation}}
\newcommand{\beqa}{\begin{eqnarray}}
\newcommand{\eeqa}{\end{eqnarray}}
\newcommand{\beqan}{\begin{eqnarray*}}
\newcommand{\eeqan}{\end{eqnarray*}}
\newcommand{\bealn}{\begin{align*}}
\newcommand{\eealn}{\end{align*}}

\newcommand{\bA}{\mathbb{A}}

\newcommand{\bI}{\mathbb{I}}

\newcommand{\bS}{\mathbb{S}}
\newcommand{\bT}{\mathbb{T}}

\newcommand{\cA}{\mathcal{A}}
\newcommand{\cB}{\mathcal{B}}
\newcommand{\cC}{\mathcal{C}}

\newcommand{\cM}{\mathcal{M}}

\newcommand{\cP}{\mathcal{P}}

\newcommand{\cS}{\mathcal{S}}

\newcommand{\cU}{\mathcal{U}}
\newcommand{\cV}{\mathcal{V}}

\newcommand{\cX}{\mathcal{X}}

\newcommand{\kR}{\mathfrak{R}}

\newcommand{\Real}{\mathbb{R}}
\newcommand{\Nat}{\mathbb{N}}

\newcommand{\Span}{\bS}
\newcommand{\Spanstar}{\Psi}
\newcommand{\Esp}{\mathbb{E}}
\newcommand{\EE}{\Esp}
\newcommand{\Var}{\mathbb{V}}
\newcommand{\var}{\mathbb{V}}

\newcommand{\KL}{\texttt{KL}}
\newcommand{\ind}{\mathbb{I}}

\renewcommand{\phi}{\varphi}
\renewcommand{\epsilon}{\varepsilon}
\renewcommand{\leq}{\leqslant}
\renewcommand{\le}{\leqslant}
\renewcommand{\geq}{\geqslant}
\renewcommand{\ge}{\geqslant}
\newcommand{\eqdef}{\stackrel{\rm def}{=}}
\newcommand{\Argmax}{\mathop{\mathrm{Argmax}}}
\usepackage{changepage}

\def\Ucrl{\texttt{\textsc{Ucrl2}}}
\newcommand{\UCRL}{\textsc{\texttt{Ucrl2}}}
\newcommand{\KLUCRL}{\textsc{\texttt{KL-Ucrl}}}

\jmlrheading{volume}{year}{pages}{submitted}{published}{meila00a}{M.~Sadegh Talebi and Odalric-Ambrym Maillard}

\ShortHeadings{Variance-Aware Regret Bounds in MDPs}{Talebi and Maillard}
\firstpageno{1}

\begin{document}

\title{Variance-Aware Regret Bounds for Undiscounted Reinforcement Learning in MDPs}

\author{\name Mohammad Sadegh Talebi\thanks{The authors contributed equally.}  \email mstms@kth.se \\
       \addr KTH Royal Institute of Technology, Stockholm, Sweden
       \AND
       \name Odalric-Ambrym Maillard\thanks{The authors contributed equally.}
        \email odalric.maillard@inria.fr \\
       \addr INRIA Lille -- Nord Europe, Villeneuve d'Ascq, France}

\editor{Mehryar Mohri and Karthik Sridharan}

\maketitle


\vspace{-3mm}
\begin{abstract}
The problem of reinforcement learning in an unknown and discrete Markov Decision Process (MDP) under the average-reward criterion is considered, when the learner interacts with the system in a single stream of observations, starting from an initial state without any reset. We revisit the minimax lower bound for that problem by making appear the local variance of the bias function  in place of the diameter of the MDP. Furthermore, we provide a novel analysis of the \KLUCRL\ algorithm establishing a high-probability regret bound scaling as  $\widetilde {\Ocal}\Bigl({\textstyle \sqrt{S\sum_{s,a}{\bf V}^\star_{s,a}T}}\Big)$ for this algorithm for ergodic MDPs, where $S$ denotes the number of states and
	where ${\bf V}^\star_{s,a}$ is the variance of the bias function with respect to the next-state distribution following action $a$ in state $s$.  The resulting bound improves upon the best previously known regret bound $\widetilde {\Ocal}(DS\sqrt{AT})$ for that algorithm, where $A$ and $D$ respectively denote the maximum number of actions (per state) and the diameter of MDP. We finally compare the leading terms of the two bounds in some benchmark MDPs indicating that the derived bound can provide an order of magnitude improvement in some cases. Our analysis leverages novel variations of the transportation lemma combined with Kullback-Leibler concentration inequalities, that we believe to be of independent interest.
\end{abstract}


\section{Introduction}
In this paper, we consider  Reinforcement Learning (RL) in an unknown and discrete Markov Decision Process (MDP) under the average-reward criterion, when the learner interacts with the system in a single stream of observations, starting from an initial state without any reset.
More formally,  let $M=(\Scal, \Acal, \nu, P)$ denote an MDP where $\Scal$ is a finite set of states and $\Acal$ is a finite set of actions available at any state, with respective cardinalities $S$ and $A$.
The reward function and the transition kernel is respectively denoted by $\nu$ and $P$.  
The game goes as follows: the learner starts in some state $s_1\in\cS$ at time $t=1$.
At each time step $t\in\Nat$, the learner chooses one action $a\in\cA$ in her current state $s\in\cS$ based on her past decisions and observations. When executing action $a$ in state $s$, the learner receives a random reward $r$ drawn independently from distribution $\nu(s,a)$ with support $[0,1]$ and mean $\mu(s,a)$. 
The state then transits to a next state $s'\in \Scal$ sampled with probability $p(s'|s,a)$, and a new decision step begins.
As the transition probabilities and reward functions are unknown, the learner has to learn them by trying different actions and recording the realized rewards and state transitions.  
We refer to standard textbooks \citep{sutton1998reinforcement,puterman2014markov} for background material on RL and MDPs.

The performance of the learner can be quantified through the notion of regret, which compares the reward collected by the learner (or the algorithm) to that obtained by an oracle always following an optimal policy, where a policy is a mapping  from states to  actions. More formally, let $\pi:\cS\to\cP(\cA)$ denote a possibly stochastic policy.
We further introduce the notation $p(s'|s,\pi(s)) = \Esp_{Z\sim \pi(s)}[p(s'|s,Z)]$,
and $P_\pi f$ to denote the function $s\mapsto\sum_{s'\in\cS} p(s'|s,\pi(s))f(s')$.
Likewise, let $\mu_\pi(s) = \Esp_{Z\sim \pi(s)}[\mu(s,Z)]$ denote the mean reward after choosing action $\pi(s)$ in step $s$.

\vspace{-3mm}
\begin{definition}[Expected cumulative reward]\label{def:cumrew}
	The expected cumulative reward of policy $\pi$  when run for $T$ steps from initial state $s_1$ is defined as
	
	\vspace{-7mm}
	\beqan
	R_{\pi,T}(s_1)=\Esp\bigg[\sum_{t=1}^T r(s_t,a_t)\bigg] = \mu_\pi(s_1) + (P_\pi\mu_\pi)(s_1) + \dots = \sum_{t=1}^T (P_\pi^{t-1}\mu_\pi)(s_1)\,.
	\eeqan
	
	\vspace{-4mm}\noindent
	where $a_t \sim \pi(s_t)$, $s_{t+1} \sim p(\cdot|s_t,a_t)$,
	and finally $r(s,a) \sim \nu(s,a)$.
\end{definition}

\vspace{-6mm}
\begin{definition}[Average gain and bias]\label{def:biasgain}
	Let us introduce the average transition operator
	$\overline{P}_\pi = \lim_{T\to\infty} \frac{1}{T}\sum_{t=1}^T P_\pi^{t-1}$.
	The average gain $g_\pi$ and bias function $b_\pi$ are defined by
	
	\vspace{-6mm}
	\beqan
	g_\pi(s_1) =
	\lim_{T\to\infty} \frac{1}{T}R_{\pi,T}(s_1)= (\overline{P}_\pi \mu_\pi)(s_1)\,,
	\qquad
	b_\pi(s) =\sum_{t=1}^\infty \Big((P_\pi^{t-1}- \overline{P}_\pi)\mu_\pi\Big)(s)\,.
	\eeqan	
\end{definition}

\vspace{-4mm}\noindent
The previous definition requires some mild assumption on the MDP for the limits to makes sense. It is shown (see, e.g., \citep{puterman2014markov}) that the average gain achieved by executing a stationary policy $\pi$ in a communicating MDP $M$ is well-defined and further does not depend on the initial state, i.e., $g_\pi(s_1)=g_\pi$. For this reason, we restrict our attention to such MDPs in the rest of this paper.
Furthermore, let $\star$ denote an optimal policy, that is\footnote{The maximum is reached since there are only finitely many deterministic policies.} $g_\star = \max_{\pi} g_\pi.$

\vspace{-2mm}
\begin{definition}[Regret]\label{def:regret}
We define the regret of any learning algorithm $\bA$ after $T$ steps as

\vspace{-7mm}
\beqan
\mathrm{Regret}_{\bA,T} := \sum_{t=1}^T r(s_t^\star,\star(s_t^\star))- \sum_{t=1}^T r(s_t,a_t)\quad
\text{where } a_t =\bA(s_t, (\{s_{t'},a_{t'},r_{t'}\})_{t'<t} )\,,
\eeqan

\vspace{-3mm}\noindent
and $s^\star_{t+1} \sim p(\cdot |s^\star_t,\star(s^\star_t))$ with $s^\star_1=s_1$ is a sequence  generated by the optimal strategy.
\end{definition}

\vspace{-3mm}\noindent
By an application of Azuma-Hoeffding's inequality for bounded  random martingales, it is immediate to show that with probability higher than $1-\delta$,

\vspace{-7mm}
\beqan
\mathrm{Regret}_{\bA,T}  &\leq& \sum_{t=1}^{T}\Big(P_\star^{t-1}	\mu_\star - P_{a_t}^{t-1} \mu_{a_t}\Big) + \sqrt{2T\log(2/\delta)}\\
&=&\sum_{t=1}^T (P_\star^{t-1}- \overline{P}_\star)	\mu_\star  + \bigg[Tg_\star -
\sum_{t=1}^T P_{a_t}^{t-1} \mu_{a_t}\bigg]+ \sqrt{2T\log(2/\delta)}\,.
\eeqan

\vspace{-4mm}\noindent
Thus, following \citep{jaksch2010near}, it makes sense to focus on the control of the middle term in brackets only. This leads us to consider the following notion of regret, which we may call \emph{effective regret}:
\begin{align*}
\kR_{\bA, T} := Tg^\star - \sum_{t=1}^T r(s_t, a_t)\, .
\end{align*}

To date, several algorithms have been proposed in order to minimize the regret based on the \emph{optimism in the face of uncertainty} principle, coming from the
literature on stochastic multi-armed bandits (see \citep{robbins1952some}).
Algorithms designed based on this principle typically maintain confidence bounds on the unknown reward and transition distributions, and choose an optimistic model that leads to the highest average long-term reward.
One of the first algorithms based on this principle for MDPs is due to \citep{burnetas1997optimal}, which is shown to be asymptotically optimal. Their proposed algorithm uses the Kullback-Leibler (KL) divergence to define confidence bounds for transition probabilities. Subsequent studies by \citep{tewari2008optimistic}, \citep{auer2007logarithmic}, \citep{jaksch2010near}, and \citep{bartlett2009regal} propose algorithms that maintain confidence bounds on transition kernel defined by $L_1$ or total variation norm. The use of $L_1$ norm, instead of KL-divergence, allows one to describe the uncertainty of the transition kernel by a polytope, which in turn brings computational advantages and ease in the regret analysis. On the other hand, such polytopic models are typically known to
 provide poor representations of underlying uncertainties; we refer to the literature on the robust control of MDPs with uncertain transition kernels, e.g., \citep{nilim2005robust}, and more appropriately to \citep{filippi2010optimism}.
 Indeed, as argued in \citep{filippi2010optimism}, optimistic models designed by $L_1$ norm suffer from two shortcomings: (i) the $L_1$ optimistic model could lead to inconsistent models by assigning a zero mass to an already observed element, and (ii) due to polytopic shape of $L_1$-induced confidence bounds, the maximizer of a linear optimization over $L_1$ ball could significantly vary for a small change in the value function, thus resulting in sub-optimal exploration (we refer to the discussion and illustrations on pages 120--121 in \citep{filippi2010optimism}).

Both of these shortcomings are avoided by making use of the KL-divergence and properties of the corresponding KL-ball.
In \citep{filippi2010optimism}, the authors introduce the \KLUCRL\ algorithm that modifies the \texttt{\textsc{Ucrl2}} algorithm of \citep{jaksch2010near} by replacing $L_1$ norms with KL divergences
in order to  define the confidence bound on transition probabilities. Further, they provide an efficient way to carry out linear optimization over the KL-ball, which is necessary in each iteration of the Extended Value Iteration.
Despite these favorable properties and the strictly superior performance in numerical experiments (even for very short time horizons), the best known regret bound for \KLUCRL\ matches that of \texttt{\textsc{Ucrl2}}. Hence, from a theoretical perspective, the potential gain of use of KL-divergence to define confidence bounds for transition function has remained largely unexplored. The goal of this paper is to investigate this gap.

\vspace{-3mm}
\paragraph{Main contributions.} In this paper we provide a new regret bound for \KLUCRL\ scaling as $\widetilde \Ocal\bigl(\sqrt{S\sum_{s,a}\mathbf V_{s,a}^\star T} + D\sqrt{T}\big)$ for ergodic MDPs with $S$ states, $A$ actions, and diameter $D$. Here, $\mathbf V^\star_{s,a}:=\Var_{p(\cdot|s,a)}(b^\star)$ denotes the variance of the optimal bias function $b^\star$ of the true (unknown) MDP with respect to next state distribution under state-action $(s,a)$. This bound improves over the best previous bound of $\widetilde \Ocal(DS\sqrt{AT})$ for \KLUCRL\ as $\sqrt{\mathbf V_{s,a}^\star}\leq D$. Interestingly, in several examples $\sqrt{\mathbf V_{s,a}^\star}\ll D$ and actually $\sqrt{\mathbf V^\star_{s,a}}$ is comparable to $\sqrt{D}$. Our numerical experiments on typical MDPs further confirm that $\sqrt{S\sum_{s,a} \mathbf V_{s,a}^\star}$ could be orders of magnitude smaller than $DS\sqrt{A}$.
To prove this result, we provide novel transportation concentration inequalities inspired by the transportation method that relate the so-called transportation cost under two discrete probability measures to the KL-divergence between the two measures and the associated variances. To the best of our knowledge, these inequalities are new and of independent interest. To complete our result, we provide a new minimax regret lower bound of order $\Omega(\sqrt{SA\mathbf V_{\max} T})$, where $\mathbf V_{\max}:=\max_{s,a} \mathbf V^\star_{s,a}$.  In view of the new minimax lower bound, the reported regret bound for \KLUCRL\ can be improved by only a factor $\sqrt{S}$. 

\vspace{-3mm}
\paragraph{Related work.}
RL in unknown MDPs under average-reward criterion dates back to the seminal papers by \citep{graves1997asymptotically}, and \citep{burnetas1997optimal}, followed by \citep{tewari2008optimistic}. Among these studies, for the case of ergodic MDPs, \citep{burnetas1997optimal} derive an asymptotic MDP-dependent lower bound on the regret and provide an asymptotically optimal algorithm.
Algorithms with finite-time regret guarantees and for wider class of MDPs are presented by \citep{auer2007logarithmic}, \citep{jaksch2010near,auer2009near}, \citep{bartlett2009regal}, \citep{filippi2010optimism}, and \citep{maillard2014hard}.

\Ucrl\ and \KLUCRL\ achieve a $\widetilde\Ocal(DS\sqrt{AT})$
regret bound with high probability in communicating MDPs, for any unknown time horizon.  \textsc{\texttt{Regal}} obtains a $\widetilde\Ocal(D' S\sqrt{AT})$ regret with high probability in the larger class of weakly communicating MDPs, provided that we know an upper bound  $D'$ on the span of the bias function. It is however still an open problem to incorporate this knowledge into an implementable algorithm.
 The \texttt{TSDE} algorithm by Ouyang et al.~\citep{ouyang2017learning} achieves a regret growing as $\widetilde \Ocal(D' S\sqrt{AT})$ for the class of weakly communicating MDPs, where $D'$ is a given bound on the span of the bias function.
In a recent study, \citep{agrawal2017posterior} propose an algorithm based on posterior sampling for the class of communicating MDPs.
Under the assumption of known reward function and \emph{known time horizon}, their algorithm enjoys a regret bound scaling as $\widetilde{\cal O}(D\sqrt{SAT})$, which constitutes the best known regret upper bound for learning in communicating MDPs and has a tight dependencies on $S$ and $A$.

We finally mention that some studies consider regret minimization in MDPs in the \textit{episodic} setting, where the length of each episode is fixed and known; see, e.g., \citep{osband2013more}, \citep{azar2017minimax}, and \citep{dann2017unifying}. Although RL in the episodic setting bears some similarities to the average-reward setting, the techniques developed in these paper strongly rely on the fixed length of the episode, which is assumed to be small, and do not directly carry over to the case of undiscounted RL considered here.

\section{Background Material and The \KLUCRL\ Algorithm}

In this section, we recall some basic material on undiscounted MDPs and then detail the \KLUCRL\ algorithm.

\vspace{-2mm}
\begin{lemma}[Bias and Gain]\label{lem:biasgain} The gain and bias function satisfy the following relations
	\beqan
	(\text{Bellman equation})\qquad b_\pi + g_\pi &=&  \mu_\pi + P_\pi b_\pi\\
		(\text{Fundamental matrix})\qquad\quad b_\pi &=& [I-P_\pi + \overline{P_\pi}]^{-1}[I- \overline{P}_\pi] \mu_\pi\,.
	\eeqan
\end{lemma}

\vspace{-1mm}\noindent
This result is an easy consequence of the fact that ${\overline P}_\pi$ (see Definition~\ref{def:biasgain}) satisfies $\overline{P}_\pi P_\pi =  P_\pi \overline{P}_\pi =
\overline{P}_\pi \overline{P}_\pi = \overline{P}_\pi\,$ (see \citep{puterman2014markov} as well as Appendix~\ref{app:background} for details).

According to the standard terminology, we say a policy is $b_\star$-improving if it satisfies
$\pi(s) = \argmax_{a\in\cA} \mu(s,a) + (P_ab_\star)(s)$\,.
Applying the theory of MDPs (see, e.g., \citep{puterman2014markov}), it can be shown that
any $b_\star$-improving policy is optimal and thus that we can choose $\star$ to satisfy\footnote{The solution to this fixed-point equation is defined only up to an additive constant. Some people tend to use this equation in order to define $b_\star$ and $g_\star$, but this is a bad habit that we avoid here.} the following fundamental identity\footnote{Throughout this paper, we may use $g^\star$ (resp.~$b^\star$) and $g_\star$ (resp.~$b_\star$) interchangeably.}
\begin{align*}
(\text{Bellman optimality equation})\quad
\forall s\in\cS,\,
b_\star(s) + g_\star  = \max_{a\in \Acal} \Bigl( \mu(s,a) + \sum_{y\in \Scal} p(y|s,a)b_\star(y) \Big)\; .
\end{align*}

\vspace{-3mm}\noindent
We now recall the definition of diameter and mixing time as we consider only MDPs with finite diameter or mixing time.
\begin{definition}[Diameter \citep{jaksch2010near}]
	Let $T_\pi(s'|s)$ denote the first hitting time of state $s'$ when following stationary policy $\pi$ from initial state $s$. The diameter $D$ of an MDP $M$ is defined as
	$$
	D := \max_{s\neq s'} \min_\pi \EE[T_\pi(s'|s)].
	$$
\end{definition}

\begin{definition}[Mixing time \citep{auer2007logarithmic}]
 \label{def:mixing_time}
 Let $\cC_\pi$ denote the Markov chain induced by the policy $\pi$ in an ergodic MDP $M$ and let $T_{\cC_\pi}$ represent the hitting time of $\cC_\pi$. The mixing time $T_M$ of $M$ is defined as
\begin{align*}
T_M:=\max_\pi T_{\cC_\pi}\, .
\end{align*}
\end{definition}

For convenience,  we also introduce,  for any function $f$ defined on $\Scal$, its span defined by $\Span(f):=\max_{s\in \Scal} f(s) - \min_{s\in \Scal} f(s)$. It actually acts as a semi-norm (see \citep{puterman2014markov}).

We finally introduce the following quantity that appears in the known problem-dependent lower-bounds on the regret, and plays the analogue of the mean gap in the bandit literature.

\vspace{-2mm}
\begin{definition}[Sub-optimality gap]
The sub-optimality of action $a$ at state $s$ is
\begin{align}
\label{eq:gap_s_a}
\phi(s,a) &= \mu(s,\star(s)) - \mu(s,a) + (p(\cdot|s,\star(s)) - p(\cdot|s,a))^\top {b_\star}\; .
\end{align}
\end{definition}

\vspace{-3mm}\noindent
 Note importantly that $\phi$ is defined in terms of the bias $b_\star$ of the optimal policy $\star$.
Indeed, it can be shown that minimizing the effective regret (in expectation) is essentially equivalent to minimizing the quantity $\sum_{s,a} \phi(s,a) \Esp[N_T(s,a)]$, where $N_T(s,a)$ is the total number of steps when action $a$ has been played in state $s$. More precisely, it is not difficult to show  (see Appendix~\ref{app:background} for completeness) that for any stationary policy $\pi$ and all $t$,

\vspace{-6mm}
\beqa\label{eq:effectiveTopseudo}
\EE[\kR_{\pi,t}] = \sum_{s,a} \phi(s,a) \Esp[N_t(s,a)] + {\big((P^{t-1}_\pi-I)b_\star\big)(s_1)}
\leq \sum_{s,a} \phi(s,a) \Esp[N_t(s,a)] + D\,.
\eeqa

\vspace{-2mm}
\paragraph{The \KLUCRL\ algorithm.}
The \KLUCRL\ algorithm \citep{filippi2010optimism, filippi2010strategies} is a model-based algorithm inspired by \texttt{\textsc{Ucrl2}} \citep{jaksch2010near}.
To present the algorithm, we first describe how it defines, at each given time $t$, the set of plausible MDPs based on the observation available at that time.
To this end, we introduce the following notations. Under a given algorithm and for a state-action pair $(s,a)$, let $N_t(s,a)$ denote the number of visits, up to time $t$, to $(s,a)$: $N_t(s,a) = \sum_{t'=0}^{t-1} \bI \{s_{t'} = s, a_{t'} = a\}$. Then, let
$N_t(s,a)^+ =\max\{N_t(s,a),1\}$. Similarly, $N_t(s,a,s')$ denotes the number of visits to  $(s,a)$, up to time $t$, followed by a visit to state $s'$: $N_t(s,a,s') = \sum_{t'=0}^{t-1} \bI \{s_{t'} = s, a_{t'} = a, s_{t'+1}=s'\}$. We introduce the empirical estimates of transition probabilities and rewards:
\begin{align*}
\hat\mu_t(s,a) = \frac{\sum_{t'=0}^{t-1} r_t \bI \{s_{t'} = s, a_{t'} = a\}}{N_t(s,a)^+}, \quad \hat p_t(s'|s,a) = \frac{N_t(s,a,s')}{N_t(s,a)^+}\,.
\end{align*}

\KLUCRL, as an optimistic model-based approach, considers the set $\Mcal_t$ as a collection of all MDPs $M'=(\Scal,\Acal, \nu', P')$, whose transition kernels and reward functions satisfy:
\beqa
  \kl(\hat{p}_t(\cdot|s,a),p'(\cdot|s,a))&\leq & C_p/ N_t(s,a)\, ,\label{eq:klucrl_conf_p}\\
   |\hat \mu_t(s,a) -  {\mu'}(s,a)|&\leq&  \sqrt{C_\mu/N_t(s,a)}\, ,   \label{eq:klucrl_conf_r}
\eeqa
where $\mu'$ denotes the mean of $\nu'$, and where $C_p:=C_p(T,\delta) = S\left(B+\log(G)(1+1/G)\right)$, with  $B=B(T,\delta):=\log(2eS^2A\log(T)/\delta)$ and $G=B+1/\log(T)$, and
$C_\mu := C_\mu(T,\delta) = \log(4SA\log(T)/\delta)/1.99$.
Importantly, as proven in \citep[Proposition~1]{filippi2010optimism}, with probability at least $1-2\delta$, the true MDP $M$ belongs to the set $\Mcal_t$ uniformly over all time steps $t\leq T$.

Similarly to \texttt{\textsc{Ucrl2}}, \KLUCRL\ proceeds in episodes of varying lengths; see Algorithm~\ref{alg:klucrl}. We index an episode by $k\in \NN$. The starting time of the $k$-th episode is denoted $t_k$, and by a slight abuse of notation, let $\Mcal_k:= \Mcal_{t_k}$, $N_k:=N_{t_k}$, $\hat\mu_k= \hat\mu_{t_k}$, and $\hat p_k:=\hat p_{t_k}$. At $t=t_k$, the algorithm forms the set of plausible MDPs $\Mcal_k$ based on the observations gathered so far.
It then defines an extended MDP $M_{\texttt{ext},k} = (\Scal, \Acal\times \cM_k,\mu_{\texttt{ext}},  P_{\texttt{ext}})$, where for an extended action $a_{\texttt{ext}}=(a,M')$, it defines
$\mu_{\texttt{ext}}(s,a_{\texttt{ext}}) = \mu'(s,a)$ and
	$p_{\texttt{ext}}(s'|s,a_{\texttt{ext}}) = p'(s'|s,a)$.
Then, a $\frac{1}{\sqrt{t_k}}$-optimal extended policy $\pi_{\texttt{ext},k}$ is computed in the form $\pi_{\texttt{ext},k}(s) = (\tilde M_k, \tilde \pi_k(s))$, in the sense that it satisfies

\vspace{-6mm}
$$
\tilde g_{k} \eqdef g_{\tilde \pi_k}(\tilde M_k) \geq  \max_{M'\in \Mcal_k, \pi} g_\pi(M') - \frac{1}{\sqrt{t_k}}\, ,
$$

\vspace{-2mm}\noindent
where  $g_\pi(M)$ denotes the gain of policy $\pi$ in MDP $M$.
$\tilde M_k$ and $\tilde \pi_k$ are respectively called the optimistic MDP and the optimistic policy.
Finally, an episode stops at the first step $t=t_{k+1}$ when the number of local counts $v_{k,t}(s,a) = \sum_{t'=t_k}^{t}\ind\{s_{t'}=s,a_{t'}=a\}$ exceeds $N_{t_k}(s,a)$ for some $(s,a)$.
We denote with some abuse $v_{k} = v_{k,t_{k+1}-1}$.

\begin{remark}
\label{rem:comptu_limit}
	The value $1/\sqrt{t_k}$ is a parameter of Extended Value Iteration and is only here for computational reasons: with sufficient computational power, it could be replaced with $0$.
\end{remark}

\vspace{-3mm}
\begin{algorithm}[h]
   \caption{\KLUCRL\ \citep{filippi2010optimism}, with input parameter $\delta\in (0,1]$ }
   \label{alg:klucrl}
   \footnotesize
\begin{algorithmic}
   \STATE \textbf{Initialize:} For all $(s,a)$, set $N_0(s,a)=0$ and $v_0(s,a)=0$. Set $t=1$, $k=1$, and observe initial state $s_1$ \vspace{1mm}
   \FOR{episodes $k\geq  1$}
       \STATE Set $t_k = t$ \vspace{1mm}
       \STATE Set $N_k(s,a) = N_{k-1}(s,a)+ v_{k-1}(s,a)$ for all $(s,a)$ \vspace{1mm}
       \STATE Find a $\tfrac{1}{\sqrt{t_k}}$-optimal policy $\tilde \pi_k$ and an optimistic MDP $\tilde M_k\in \Mcal_k$ using \textsc{Extended Value Iteration}
       \WHILE{$v_{k}(s_t,a_t)\geq  N_{k}(s_t,a_t)$}
            \STATE Play action $a_t=\tilde \pi_k(s_t)$, and observe the next state $s_{t+1}$ and reward $r(s_t, a_t)$ \vspace{1mm}
            \STATE Update $N_{k}(s,a,x)$ and $v_{k}(s,a)$ for all actions $a$ and states $s,x$ \vspace{1mm}
       \ENDWHILE
   \ENDFOR
\end{algorithmic}
\normalsize
\end{algorithm}

\vspace{-2mm}
\section{Regret Lower Bound}
In order to motivate the dependence of the regret on the local variance,
we first provide the following  minimax lower bound that makes appear this scaling.
\begin{theorem}
\label{thm:minimax_LB}
There exists an MDP $M$ with $S$ states and $A$ actions with $S,A\geq  10$, such that the expected regret under any algorithm $\bA$ after $T\geq  DSA$ steps for any initial state satisfies
\begin{align*}
\EE[\mathfrak{R}_{\bA,T}] \geq  0.0123\sqrt{\mathbf V_{\max} SAT}, \qquad\text{where }\qquad \mathbf V_{\max} := \max_{s,a} \Var_{p(\cdot|s,a)}(b^\star)\,.
\end{align*}
\end{theorem}

Let us recall that \citep{jaksch2010near} present a minimax lower bound on the regret scaling as $\Omega(\sqrt{DSAT})$. Their lower bound follows by considering a family of \emph{hard-to-learn} MDPs. To prove the above theorem, we also consider the same MDP instances as in \citep{jaksch2010near} and leverage their techniques. We however show that choosing a slightly different choice of transition probabilities for the problem instance leads to a lower bound scaling as $\Omega(\sqrt{\mathbf V_{\max} SAT})$, which does not depend on the diameter (the details are provided in the appendix).

We also remark that for the considered problem instance, easy calculations show that for any state-action pair $(s,a)$, the variance of bias function satisfies $c_1\sqrt{D}\leq \Var_{p(\cdot|s,a)}(b^\star) \leq c_2D$ for some constants $c_1$ and $c_2$. Hence, the lower bound in Theorem \ref{thm:minimax_LB} can serve as an alternative minimax lower bound without any dependence on the diameter.

\section{Concentration Inequalities and The Kullback-Leibler Divergence}
Before providing the novel regret bound for the \KLUCRL\ algorithm, let us discuss some important tools that we use for the regret analysis. We believe that these results, which  could also be of independent interest beyond RL, shed light on some of the challenges of the regret analysis.

Let us first recall a powerful result from mathematical statistics  (we provide the proof in Appendix~\ref{app:concentration} for completeness) known as the transportation lemma; see, e.g., \citep[Lemma~4.18]{boucheron2013concentration}:

\vspace{-2mm}
\begin{lemma}[Transportation lemma]\label{lem:transportation}
	For any function $f$, let us introduce $\phi_f:\lambda \mapsto \log \Esp_P [\exp (\lambda (f(X) - \Esp_P[f]))]$.
	Whenever $\phi_f$ is defined on some possibly unbounded interval $I$ containing $0$, define its dual	$\phi_{\star,f}(x) = \sup_{\lambda \in I} \lambda x - \phi_f(\lambda)$. Then it holds
	\beqan
	\forall Q \ll P,\quad
	\Esp_{Q}[f]-\Esp_P[f] &\leq& \phi_{+,f}^{-1}(\emph{\KL}(Q,P))\quad\text{where\; }\,\phi_{+,f}^{-1}(t) = \inf \{x \geq  0 : \phi_{\star,f}(x)>t\}\, ,\\
	\forall Q \ll P,\quad
	\Esp_{Q}[f]-\Esp_P[f] &\geq& \phi_{-,f}^{-1}(\emph{\KL}(Q,P))\quad\text{where\; }\, \phi_{-,f}^{-1}(t) = \sup \{x \leq 0 : \phi_{\star,f}(x)>t\}\,.
	\eeqan
\end{lemma}

This result is especially interesting when $Q$ is the empirical version of $P$ built from $n$ i.i.d.~observations, since in that case  it enables to \textit{decouple} the concentration properties of the distribution from the specific structure of the considered function. Further, it shows that controlling the KL divergence between $Q$ and $P$ induces a concentration result valid for all (nice enough) functions $f$, which is especially useful when we do not know in advance the function $f$ we want to handle (such as bias function $b_\star$).

The quantities  $\phi_{+,f}^{-1}$, $\phi_{-,f}^{-1}$ may look complicated. When $f(X)$ (where $X\sim P$)  is Gaussian, they coincide with $t\mapsto\pm\sqrt{2\Var_P(f)t}$. Controlling them in general is challenging. However for bounded functions, a Bernstein-type relaxation can be derived that uses the variance $\Var_P(f)$ and the span $\bS(f)$:

\vspace{-2mm}
\begin{corollary}[Bernstein transportation]\label{cor:transportationII}
For any function $f$ such that  $\Var_P[f]$ and $\bS(f)$ are finite,

\vspace{-7mm}
	\beqan
	\forall Q \ll P,\quad
	\Esp_{Q}[f]-\Esp_P[f] &\leq&\sqrt{ 2\Var_P[f]\emph{\KL}(Q,P)}  +\frac{2}{3} \bS(f)\emph{\KL}(Q,P)\,,\\
		\forall Q \ll P,\quad
	\Esp_{P}[f]-\Esp_Q[f] &\leq&\sqrt{ 2\Var_P[f]\emph{\KL}(Q,P)}\,.
	\eeqan		
\end{corollary}

We also provide below another variation of this result that is especially useful when the bounds of Corollary~\ref{cor:transportationII} cannot be handled, and that seems to be new (up to our knowledge):

\vspace{-2mm}
\begin{lemma}[Transportation method II]\label{lem:transportationII}
	\label{lem:KL_var}Let $P\in\cP(\cX)$  be a probability distribution on a finite alphabet $\cX$. Then, for any real-valued function $f$ defined on $\cX$, it holds that
\beqan
	\forall P\ll Q,\quad\Esp_Q[f] - \Esp_P[f] &\leq& \Bigl(\sqrt{\cV_{P,Q}(f)}
	+ \sqrt{\cV_{Q,P}(f)} \Big)\sqrt{2\emph{\KL}(P,Q)}+ \Span(f)\emph{\KL}(P,Q)\,,\\
	\text{where}&&
	\cV_{P,Q}(f) \,:=\sum_{x\in \cX: P(x)\geq Q(x)}\!\!\!P(x)(f(x) - \Esp_P[f])^2\,.
\eeqan
\end{lemma}

\vspace{-2mm}\noindent
When $P$ is the transition law under a state-action pair $(s,a)$ and $Q$ is its empirical estimates up to time $t$, i.e.~$Q=\hat p_t(\cdot|s,a)$ and $P=p(\cdot|s,a)$, the first assertion in Corollary \ref{cor:transportationII} can be used to decouple $\Esp_{Q}[f]-\Esp_P[f]$ from specific structure of $f$. In particular, if $f$ is some bias function, then $f$ has a bounded span $D$, and since $\KL(Q,P)=\widetilde\Ocal(N_t^{-1})$, the first order terms makes appear the variance of $f$.
This would result in a term scaling as $\widetilde \Ocal(\sqrt{S\sum_{s,a}\mathbf V_{s,a}^\star T})$ in our regret bound, where $\widetilde \Ocal(\cdot)$ hides poly-logarithmic terms.

Now, for the case when $Q=\hat p_t(\cdot|s,a)$ and $P=\tilde p_t(\cdot|s,a)$ is the optimistic transition law at time $t$, the second inequality in Corollary~\ref{cor:transportationII} allows us to bound $\Esp_P[f]-\Esp_Q[f]$ by the variance of $f$ under law $\tilde p(\cdot|s,a)$, which itself is controlled by the variance of $f$ under the true law $p(\cdot|s,a)$. Using such an approach would lead to a term scaling as $\widetilde \Ocal(\sqrt{S\sum_{s,a}\mathbf V_{s,a}^\star T} + DS^{2}T^{1/4})$. We can remove the term scaling as $\widetilde\Ocal(T^{1/4})$ in our regret analysis by resorting to Lemma~\ref{lem:transportationII} instead, in combination with the following property of the operator $\Vcal$:
\begin{lemma}
\label{lem:Vcal_properties}
Consider two distributions $P,Q\in\cP(\cX)$ with $|\Xcal|\geq 2$. Then, for any real-valued function $f$ defined on $\Xcal$, it holds that
\begin{align*}
(i) \quad \Vcal_{P,Q}(f) &\leq \Var_{P}(f) \; ,\\
(ii) \quad \sqrt{\Vcal_{P, Q}(f)} &\leq  \sqrt{2\Var_{Q}(f)} + 3\Span(f)\sqrt{|\Xcal|\emph{\kl}(Q,P)}  \; .
\end{align*}
\end{lemma}
%

%
%

\section{Variance-Aware Regret Bound for \KLUCRL}
\label{sec:klucrl_analysis}
In this section, we present a regret upper bound for \KLUCRL\ that leverages the results presented in the previous section.
Let $\Spanstar:= \Span(b^\star)$ denote the span of the bias function, and for any $(s,a)\in \Scal\times \Acal$ define $\mathbf V_{s,a}^\star := \Var_{p(\cdot|s,a)}(b^\star)$ as the variance of the bias function under law $p(\cdot|s,a)$. 

Let $\tilde \star_k$ denote the optimal policy in the extended MDP $\cM_k$, whose gain $\tilde g_{\tilde \star_k}$ satisfies $\tilde g_{\tilde \star_k} = \max_{M'\in \cM_k, \pi} g_\pi(M')$. We consider a variant of \KLUCRL, which computes, in every episode $k$, a policy $\tilde \pi_k$ satisfying: $\max_s |\tilde b_k(s) - \tilde b_{\tilde \star_k}(s)| \le \frac{1}{\sqrt{t_k}}$, and $\tilde g_k \ge \tilde g_{\tilde \star_k} -\frac{1}{\sqrt{t_k}}$.\footnote{We study such a variant to facilitate the analysis and presentation of the proof. This variant of \KLUCRL\ may be computationally less efficient than Algorithm \ref{alg:klucrl}. We stress however that, in view of the number of episodes (growing as $SA\log(T)$) as well as Remark \ref{rem:comptu_limit}, with sufficient computational power such an algorithm could be practical.}

In the following theorem, we  provide a refined regret bound for \KLUCRL:

\begin{theorem}[Variance-aware regret bound for \KLUCRL]
\label{thm:klucrl_reg}
With probability at least $1-6\delta$, the regret under \KLUCRL\ for any ergodic MDP $M$ and for any initial
state satisfies
\begin{align*}
\mathfrak{R}_{\KLUCRL,T} &\leq \Bigl(31\sqrt{S\textstyle \sum_{s,a} \mathbf V^\star_{s,a}} + 35S\sqrt{A} + \sqrt{2}D+1\Big)\sqrt{TB(T,\delta)} \\
&+ \widetilde {\Ocal}\Bigl(SA(T_MSA + D + S^{3/2})\log(T)\Big) \; ,
\end{align*}
where $\widetilde {\Ocal}$ hides the terms scaling as $\mathrm{polylog}(\log(T)/\delta)$. Hence, with probability at least $1-\delta$,
$$
\mathfrak{R}_{\KLUCRL,T} = \Ocal\Bigl(\bigl[\sqrt{S\textstyle \sum_{s,a} \mathbf V^\star_{s,a}} + D\big]\sqrt{T\log(\log(T)/\delta)}\Big)\, .
$$
\end{theorem}

\begin{remark}
If the cardinality of the set $\Scal_{s,a}^+:=\{s': p(s'|s,a)>0\}$ for state-action $(s,a)$ is known, then one can use the following improved confidence bound for the pair $(s,a)$  (instead of (\ref{eq:klucrl_conf_p})):
\begin{align}
   N_t(s,a)\emph{\kl}(\hat{p}_t(\cdot|s,a),p'(\cdot|s,a))&\leq  C^{s,a}_p\;,\label{eq:klucrl_conf_p_known_support}
\end{align}
where $C^{s,a}_p = \frac{|\Scal_{s,a}^+|}{S}C_p$ (see, e.g., \citep[Proposition~4.1]{filippi2010strategies} for the corresponding concentration result). As a result, if $|\Scal_{s,a}^+|$ for all $(s,a)\in \Scal\times \Acal$ is known, it is then straightforward to show that the corresponding variant of \KLUCRL, which relies on (\ref{eq:klucrl_conf_p_known_support}), achives a regret growing as
$
\widetilde \Ocal\bigl(\sqrt{\textstyle {\sum_{s,a}|\Scal_{s,a}^+|\mathbf V_{s,a}^\star T}} + D\sqrt{T}\big).
$
\end{remark}

The regret bound provided in the aforementioned remark is of particular importance in the case of \emph{sparse MDPs},  where most states transit to only a few next-states under various actions. We would like to stress that to get an improvement of a similar flavour for \UCRL, to the best of our knowledge, one has to know the sets $\Scal_{s,a}^+$ for all $(s,a)\in \Scal\times \Acal$ rather than their cardinalities.

\paragraph{Sketch of proof of Theorem \ref{thm:klucrl_reg}.}
The detailed proof of this result is provided in Appendix~\ref{app:Reg_UB}.
In order to better understand it, we now provide a high level sketch of proof explaining the main steps of the analysis.

First note that by an application of Azuma-Hoeffding inequality, the effective regret is upper bounded by
$
\kR_{\bA,T} \leq  Tg_\star - \sum_{t=1}^T \mu(s_t,a_t) + \sqrt{T\log(1/\delta)/2} ,
$
with probability at least $1-\delta$. We proceed by decomposing the term $Tg_\star - \sum_{t=1}^T \mu(s_t,a_t)$ on the episodes $k=1,\dots, m(T)$, where $m(T)$ is the total number of episodes after $T$ steps. Introducing $v_k(s,a)$ as the number of visits to $(s,a)$ during episode $k$ for any $(s,a)$ and $k$, with probability at least $1-\delta$ we have
\beqan
\kR_{\bA,T} \leq  \sum_{k=1}^{m(T)} \Delta_{k} + \sqrt{T\log(1/\delta)/2}
= \sum_{k=1}^{m(T)}  \Delta_k\quad
\text{where } \Delta_k=\sum_{s,a} v_k(s,a)(g^\star - \mu(s,a))\,.
\eeqan
We focus on episodes such that $M\in \cM_k$, corresponding to valid confidence intervals, up to losing a probability only $2\delta$. In order to control $\Delta_k\ind\{M \in\cM_k \}$, we use the decomposition
\beqan
\sum_{s,a} v_k(s,a)(g^\star - \mu(s,a))
= \sum_{s,a} v_k(s,a)(\tilde g_k - \mu(s,a) + (g^\star - \tilde g_k)) \, .
\eeqan
We refrain from using the fact that $g^\star-  \tilde g_k \leq 1/\sqrt{t_k}$ and instead use it as a slack later in the proof. We then introduce the bias function from the identity $\tilde g_k - \tilde \mu_k = (\widetilde P_{k}-I)\tilde b_{k}$, and thus get
\beqan
\Delta_k 
&=&\sum_{s,a}v_k(s,a)\Big(\underbrace{(\widetilde P_{k}-P_{k})b_\star}_{(a)} + \underbrace{(P_{k}- I)\tilde b_{k}}_{(b)} + \underbrace{(\widetilde P_{k} - P_{k})
(\tilde b_{k}-b_\star)
+(g^\star-  \tilde g_k)}_{(c)}\Big)
\eeqan

\noindent
{\bf Term (a).}
The first term is controlled thanks to our variance-aware concentration inequalities:

\vspace{-6mm}
\beqan
(\widetilde P_{k}-P_{k})b_\star
&=&(\widehat P_{k}-P_{k})b_\star
+(\widetilde P_{k}-\widehat P_{k})b_\star\, ,\quad \text{ where}\\
\forall s,\quad ((\widehat P_{k}-P_{k})b_\star)(s)&\leq& \sqrt{2 {\bf V}^\star_{s,\tilde \pi_k(s)} \KL(\hat p_k,p)}+\frac{2}{3}\bS(b_\star) \KL(\hat p_k,p)\quad\text{ and}\\
\forall s,\quad((\widetilde P_{k}-\widehat P_{k})b_\star)(s)&\leq&(1+\sqrt{2})\sqrt{2\Var_{\hat p_k}(b_\star)\KL(\hat p_k,\tilde p_k)} + \bS(b_\star)(1+ 3\sqrt{2S}) \KL(\hat p_k,\tilde p_k)\,.
\eeqan
The first inequality is obtained by Corollary~\ref{cor:transportationII}
while the second one by a combination of Lemma~\ref{lem:transportationII}
together with Lemma~\ref{lem:Vcal_properties}. We then relate
$\sqrt{\Var_{\hat p_k}(b_\star)}$
to $\sqrt{\Var_{p}(b_\star)}$ thanks to:
\begin{lemma}
	\label{lem:Var_phat}
	For any episode $k\geq  1$ such that $M\in \Mcal_k$, it holds that for any pair $(s,a)$,
	\begin{align*}
	\sqrt{\Var_{\hat p_k(\cdot|s,a)}(f)} &\leq \sqrt{2\Var_{p(\cdot|s,a)}(f)} + \frac{6S\Span(f)B}{\sqrt{N_k(s,a)}}
	\quad\text{	with probability at least }1-\delta.
	\end{align*}
\end{lemma}
It is then not difficult to show that this first term, when summed over all episodes, contributes to the regret as
$\widetilde {\cal O}(\sqrt{S \sum_{s,a} {\bf V}^\star_{s,a}}\sqrt{T\log(\log(T)/\delta)})$,
where the $\log(\log(T))$ terms comes from the use of time-uniform concentration inequalities.

\noindent
{\bf Term (b).}
We then turn to Term (b) and observe that it makes appear a martingale difference structure. Following the same reasoning as in \citep{jaksch2010near} or \citep{filippi2010optimism}, the right way to control it is however to sum this contribution over all episodes and make appear a martingale difference sequence of $T$ deterministic terms, bounded by the deterministic quantity $D$, since $\bS(\tilde b_{k}) \leq D$. This comes at the price of losing a constant error $D$ per episode.
Now, since it can be shown that $m(T) \leq SA\log_2(8T/SA)$ as for \UCRL, we deduce that with probability higher than $1-\delta$,
\beqan
\sum_{k=1} ^{m(T)}\sum_{s,a}v_k(s,a)(P_{k}-I) \tilde b_{k} \leq
D\sqrt{2T\log(1/\delta)} + 2DSA\log_2(8T/SA)\,.
\eeqan

\noindent
{\bf Term (c).} It thus remains to handle Term (c). To this end,
we first partition the states into $\cS_s^+ = \{ x\in\cS: \widetilde P_{k}(s,x) > P_{k}(s,x)\}$ and its complementary set $\cS_s^{-}$, and get
\beqan
v_k(\widetilde P_{k} - P_{k})
(\tilde b_{k}-b_\star)
&=& \sum_s v_k(s,\tilde\pi_k(s))\sum_{x\in\cS^+_s}(\widetilde P_{k}(s,x) - P_{k}(s,x))(\tilde b_{k}(x)-b_\star(x))\\
&+& \sum_s v_k(s,\tilde\pi_k(s))\sum_{x\in\cS^-_s}(\widetilde P_{k}(s,x) - P_{k}(s,x))(\tilde b_{k}(x)-b_\star(x))\,.
\eeqan
We thus need to control the difference of bias from above and from below.
To that end, we note that by property of the bias function, it holds that
\beqan
\tilde b_{k}-b_\star = \underbrace{\tilde g_{\tilde \star_k} - \tilde g_k + (\tilde \mu_k - \mu_k) + (\widetilde P_k-P_k) b_\star}_{(d)} - \phi_{k} + \widetilde P_k (\tilde b_k-b_\star)\,.
\eeqan
Owing to the fact that $\tilde g_{\tilde\star_k} - \tilde g_k \leq 1/\sqrt{t_k}$ and by the previous results on concentration inequalities, the term (d) can be shown to be scaling as $\widetilde {\cal O}\Bigl(\sqrt{\frac{S{\bf V}^\star_{s,a}}{N_k(s,a)}}\Big)$. Thus, this means that provided that for all $s,a$, $N_k(s,a) \gtrsim \frac{S{\bf V}^\star_{s,a}}{\phi(s,a)^2}$, then
	$(d)-\phi(s,a)  \leq 0$, and thus
	$\tilde b_{k}-b_\star \leq 0+
	\widetilde P_k (0 + \ldots) \leq 0$.
On the other hand, for the control of the last term, we first note that
for an $\tilde b_{\tilde \star_k}$-improving policy (which is optimal in the extended MDP), then for all $J\in\Nat$ it  holds
\beqan
b_\star - \tilde b_{\tilde \star_k} &\leq& (\tilde g_{\tilde \star_k} - g_\star) + P_\star (b_\star - \tilde b_{\tilde \star_k}) \leq J(\tilde g_{\tilde \star_k} - g_\star) + P_\star^J (b_\star - \tilde b_{\tilde \star_k})\,.
\eeqan
Thus, we obtain that
\begin{align}
v_k(\widetilde{P}_k - P_k)(\tilde b_k &- b_\star)+ v_k(g_\star -\tilde g_{\tilde\star_k})\mathbf 1
\leq \sum_s v_k(s,\tilde\pi_k(s))\sum_{x\in \cS_s^-}(P_k(s,x)-\widetilde P_k(s,x))\bigl(P_\star^J(b_\star - \tilde b_{\tilde \star_k})\big)(x)\sk
\label{eq:ub_sketch}
&+ \sum_{s} v_k(s,\tilde\pi_k(s))\Bigl[1 - J\sum_{x\in \cS_s^-}(P_{k}(s,x)-\widetilde P_k(s,x))\Big](g_\star - \tilde g_{\tilde \star_k}) + \eta_k\,,
\end{align}
where $\eta_k$ is controlled by the error of computing $\tilde b_k$ in episode $k$ (which, for the considered variant of the algorithm, is bounded by $\sqrt{32SB}\sum_{s,a}\frac{v_k(s,a)}{N_k(s,a)^+}$). In order to handle the remaining terms in (\ref{eq:ub_sketch}), and choose $J$, we use the fact that $P_\star$ is $\gamma$-contracting for some $\gamma<1$. Thus, choosing $J = \frac{\log(D)}{\log(1/\gamma)}$ ensures that contribution of the first term in (\ref{eq:ub_sketch}) is less than $\sqrt{32SB}\sum_{s,a} \frac{v_k(s,a)}{\sqrt{N_k(s,a)^+}}$. Furthermore, provided that $N_k(s,a) \gtrsim SBJ^2$ for all $s$ and $a$, we observe that the term in brackets is non-negative, and hence the second term in (\ref{eq:ub_sketch}) becomes negative (later on we consider the case where this condition is not satisfied).
Putting together, we get $(c) \le (2\sqrt{32SB} + 1)\sum_{s,a}\frac{v_k(s,a)}{\sqrt{N_k(s,a)^+}}$.

Finally, it remains to handle the case where some state-action pair is not sufficiently sampled, that is there exists $(s,a)$ such that $N_k(s,a)< \ell_{s,a}$, where
$$
\ell_{s,a}=\ell_{s,a}(T,\delta):=\widetilde \Ocal\Bigl(SB\max\Bigl\{\frac{\Psi}{\phi(s,a)}, \frac{\log(D)}{\log(1/\gamma)}\Big\}^2\Big)\, , \quad \forall s,a.
$$ Borrowing some arguments from \citep{auer2007logarithmic}, we show that a given state-action pair $(s,a)$, which is not sufficiently sampled, contributes to the regret (until it becomes sufficiently sampled) by at most $\Ocal(T_M\max(\ell_{s,a},\log(SA/\delta)))$ with probability at least $1-\frac{\delta}{SA}$. Summing over $(s,a)$ gives the total contribution to regret.  At this point, the proof is essentially done, up to some technicalities and careful handling of second order terms.

\begin{remark}
Most steps in the proof of Theorem \ref{thm:klucrl_reg} carries over to the case of communicating MDPs without restriction (up to  considering the fact that for a communicating MDP, $P_\star$ may not induce a contractive mapping. Yet there exists some integer $\beta\ge 1$ such that $P_\star^\beta$ induces a contractive mapping; this will only affect the terms scaling as $\widetilde {\Ocal}(\log(T))$ in the regret bound). It is however not clear how to appropriately bound the regret when some state-action pair is not sufficiently sampled.
\end{remark}

\paragraph{Illustrative numerical experiments.}
In order to better highlight the magnitude of the main terms in Theorem \ref{thm:klucrl_reg} when compared to other existing results, we consider a standard class of environments for which we compute them explicitly.

\begin{figure}[t]
\centering
\footnotesize
\def\svgwidth{0.95\columnwidth}
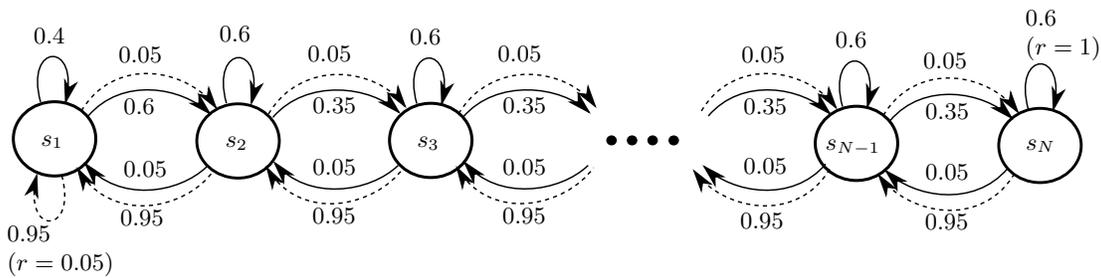
\caption{The $N$-state Ergodic \emph{RiverSwim} MDP}\label{fig:RiverSwim}
\end{figure}

For the sake of illustration, we consider the \emph{RiverSwim} MDP, introduced in \citep{strehl2008analysis}, as our benchmark environment. In order to satisfy ergodicity, here we consider a slightly modified version of the original \emph{RiverSwim} (see Figure \ref{fig:RiverSwim}). Furthermore, to convey more intuition about the potential gains, we consider varying number of states. The benefits  of \KLUCRL\ have already been studied experimentally in \citep{filippi2010optimism}, and we compute in  Table \ref{table:Span_vs_Var} features that we believe explain the reason behind this. In particular, it is apparent that while $\Spanstar\sqrt{SA}\leq D\sqrt{SA}$ grows very large as $S$ increases, $\mathbf V^\star_{s,a}$ is very small,  on all tested environments,  and does not change as $S$ increases. Further, even on this simple environment, we see that
$\sqrt{\sum_{s,a} \mathbf V^\star_{s,a}}$ is an order or magnitude smaller than
$\Spanstar\sqrt{SA}$. We believe that these computations highlight the fact that the regret bound of Theorem~\ref{thm:klucrl_reg} captures a massive improvement over the initial analysis of \KLUCRL\ in \citep{filippi2010optimism}, and over alternative algorithms such as \textsc{\texttt{Ucrl2}}.


\begin{table}
\footnotesize
\centering
\begin{tabular}[b]{cccccc}
\hline
 $S$ & $\Spanstar$ & $\max_{s,a} \mathbf V^\star_{s,a}$ &  $\Spanstar\sqrt{SA}$ & $\sqrt{\sum_{s,a} \mathbf V^\star_{s,a}}$\\ \hline
$6$ & 6.3 & 0.6322 & 21.9  &  1.8  \\
$12$ & 14.9 & 0.6327  &  72.9 & 2.8 \\
$20$ & 26.3 & 0.6327 &  166.4 & 3.7  \\
$40$ & 54.9 & 0.6327 &  490.9 & 5.3 \\
$70$ & 97.7 & 0.6327 &  1156.5 & 7.1  \\
$100$ & 140.6 & 0.6327 &  1988.3 & 8.5 \\ \hline
\end{tabular}
\caption{Comparison of span and variance for $S$-state \emph{Ergodic RiverSwim}.}
\label{table:Span_vs_Var}
\normalsize
\end{table}

\section{Conclusion}
In this paper, we revisited the analysis of  \KLUCRL\ as well as the lower bound on the regret in ergodic MDPs, in order to make appear the  local variance of the  bias function of the MDP.
Our findings show that, owing to properties of the Kullback-Leibler divergence, the leading term $\widetilde {\cal O}(DS\sqrt{AT})$ obtained for the regret  of \KLUCRL\ and \UCRL\ can be reduced to $\widetilde {\cal O}\Bigl(\sqrt{S \sum_{s,a} {\bf V}^\star_{s,a} T}\Big)$, while the lower bound for any algorithm can be shown to be $\Omega(\sqrt{SA {\bf V}_{\max} T})$, where $\mathbf V_{\max}:=\max_{s,a} \mathbf V^\star_{s,a}$.
Computations of these terms in some illustrative MDP show that the reported
upper bound may improve an order of magnitude over the existing ones (as observed experimentally in \citep{filippi2010strategies}), thus highlighting the fact that trading the diameter of the MDP to the local variance of the bias function may result in huge improvements.

We note that this improvement often corresponds to a gain of a factor ${\cal O}(\sqrt{D})$.
A natural question is whether the $\sqrt{S}$ gap between the upper and lower bounds can be filled in.
In the simpler setting of episodic reinforcement learning with known horizon $H$, several papers have shown that by taking advantage of this knowledge, it is possible to design strategies for which the regret bound does not lose a $\sqrt{S}$ factor.  However, such strategies do not apply straightforwardly to undiscounted reinforcement learning. Nonetheless, we believe that combining techniques of such studies with the tools that we have developed is a fruitful research direction.

\section*{Acknowledgment}
M.~S.~Talebi acknowledges the Ericsson Research Foundation for supporting his visit to INRIA Lille Nord -- Europe.
This work has been supported by CPER Nord-Pas de Calais/FEDER DATA Advanced data science and technologies 2015-2020,
the French Ministry of Higher Education and Research, INRIA, and the French Agence Nationale de la Recherche (ANR), under grant ANR-16- CE40-0002 (project BADASS).


\vskip 0.2in
\bibliography{bandit_RL_bib}

\appendix

\appendix

\section{Regret Lower Bound}
The proof of Theorem \ref{thm:minimax_LB} mainly relies on the problem instance for the derivation of the minimax lower bound in \citep{jaksch2010near} and related arguments there. 
For the sake of completeness, we first recall their problem instance and then compute the variance of the corresponding bias function.

\begin{figure}[h]
\centering
\def\svgwidth{0.6\columnwidth}
\small
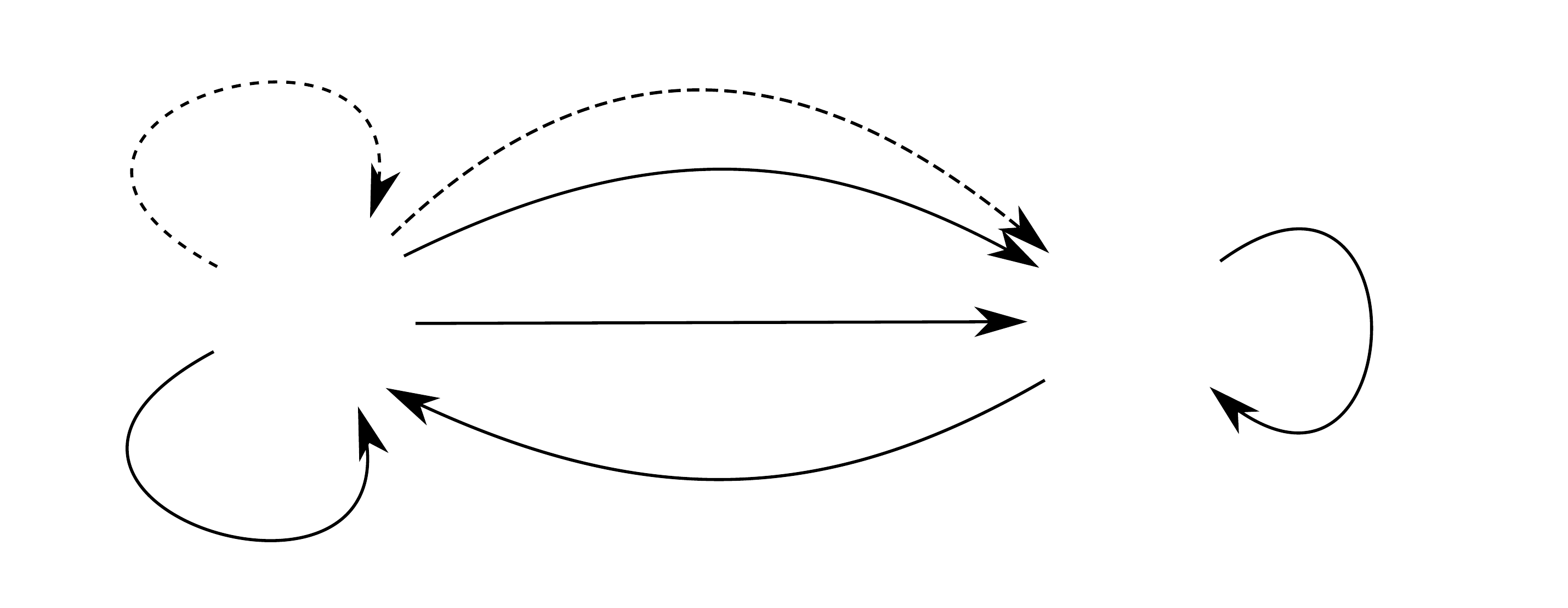
  \caption{The MDP $M'$ for lower bound \citep{jaksch2010near}}\label{fig:hard_MDP_1}
\end{figure}

To get there, we first consider the two-state MDP $M'$ shown in Figure \ref{fig:hard_MDP_1}, where there are two states $\{s_0,s_1\}$, each having $A'=\lfloor\frac{A-1}{2}\rfloor$ actions. We consider deterministic rewards defined as $r(s_0,a)=0$ and $r(s_1,a)=1$ for all $a\in \Acal$. The learner knows the rewards but not the transition probabilities. Let $\delta:=\frac{4}{D}$, where $D$ is the diameter of the MDP for which we derive the lower bound. Under any action $a$, $p(s_0|s_1,a)=\delta$. In state $s_0$, there is a unique optimal action $a^\star$, which will be referred to as the \emph{good} action. For any $a\neq a^\star$, we have $p(s_1|s_0,a)=\delta$ whereas $p(s_1|s_0,a^\star)=\delta+\varepsilon$ for some $\varepsilon\in (0,\frac{\delta}{2})$ that will be determined later. Note that the diameter $D'$ of $M'$ satisfies: $D'=\frac{1}{\delta}=\frac{D}{4}$.

We consider $\delta \in (0, {1\over 3})$.\footnote{The case of $\delta >1/3$ can be handled similarly to the analysis of \citep{jaksch2010near}.}
After straightforward calculations, one finds that   the average reward in $M'$ is given by
\begin{align*}
g^\star = \frac{1/\delta}{1/\delta + 1/(\delta+\varepsilon)} = \frac{\delta + \varepsilon}{2\delta + \varepsilon}\; .
\end{align*}
Furthermore, from Bellman optimality equation we obtain
\begin{align*}
b^\star(s_0) + \frac{\delta + \varepsilon}{2\delta + \varepsilon} = (\delta + \varepsilon) b^\star(s_1) + (1-\delta - \varepsilon) b^\star(s_0)\; ,
\end{align*}
thus giving $\Spanstar: = \Span(b^\star)=b^\star(s_1)-b^\star(s_0) = \frac{1}{2\delta+\varepsilon}$. Consider $a\neq a^\star$ and let  $p=p(\cdot|s_0,a)$. It follows that:
\begin{align*}
\EE_{p} [ b^\star] &= \delta b^\star(s_1) + (1-\delta) b^\star(s_0) = b^\star(s_0) + \delta \Spanstar\; , \\
\Var_p( b^\star) &= \delta (b^\star(s_1) - \EE_p[ b^\star])^2 + (1-\delta) (b^\star(s_0) - \EE_p[ b^\star])^2= \delta(1-\delta) \Spanstar^2\; .
\end{align*}
Similarly, we obtain
\begin{align*}
\var_{p(\cdot|s_0,a^\star)}( b^\star) &= (\delta+\varepsilon)(1-\delta-\varepsilon)\Spanstar^2\, ,\\
\var_{p(\cdot|s_1,a)}( b^\star) &= \delta(1-\delta)\Spanstar^2\, , \quad \forall a.
\end{align*}
Hence, using the facts that  $x\mapsto x(1-x)$ is increasing for $x\in [0,\tfrac{1}{2}]$ and $\varepsilon +\delta \leq \tfrac{1}{2}$, we obtain
\begin{align*}
\mathbf V_{\max} := \max_{s,a} \Var_{p(\cdot|s,a)}(b^\star) = (\delta+\varepsilon)(1-\delta-\varepsilon)\Spanstar^2\; .
\end{align*}

\subsubsection{The composite MDP} 
\begin{figure}
  \centering
  \includegraphics[scale=.55]{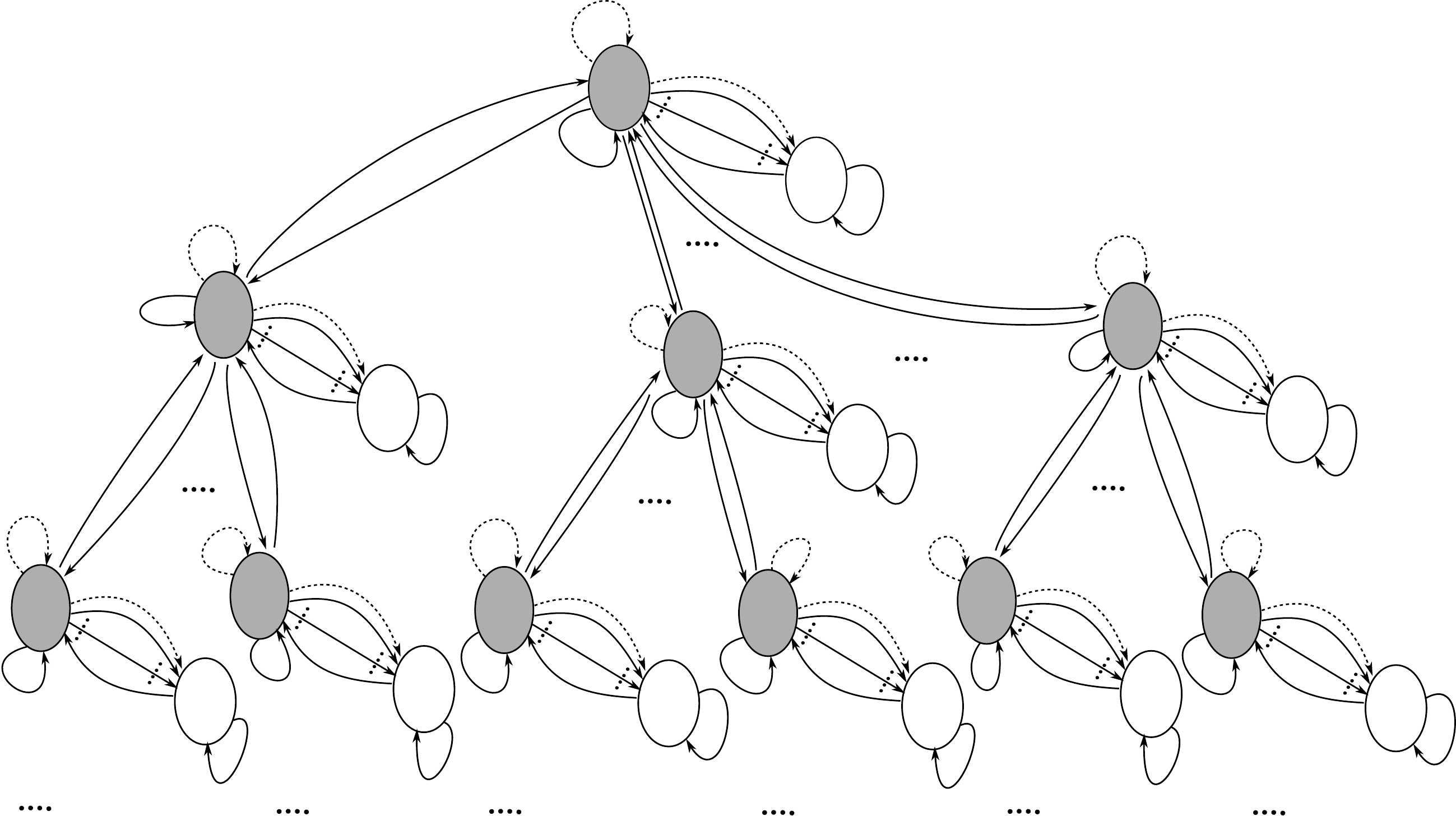}
  \caption{The composite MDP $M$ \citep{jaksch2010near}}\label{fig:hard_MDP_1_composite}
\end{figure}

We now build a composite MDP $M$ as considered in \citep{jaksch2010near}, as a concatenation of $k:=\lfloor \frac{S}{2}\rfloor$ copies of $M'$ in the form of an $A'$-ary tree, where only one copy contains the good action $a^\star$ (see Figure \ref{fig:hard_MDP_1_composite}). To this end, we first add $A'+1$ additional actions so that $M$ has at most $A$ actions per state. For any state $s_0$, one of these new actions connects $s_0$ to the root, and the rest connect $s_0$ to the leaves. Whereas for any state $s_1$, all new actions make a transition to the same state $s_1$.
By construction, the diameter of the composite MDP $M$ does not exceed $2(\frac{D}{4}+ \log_{A'} k)$, so that MDP $M$ has $2\lfloor \frac{S}{2}\rfloor \leq S$ states, $\lfloor\frac{A'-1}{2}\rfloor + \lfloor\frac{A'-1}{2}\rfloor  +1 \leq A$ actions, and a diameter less than $D$.


\subsection{Proof of Theorem \ref{thm:minimax_LB}}
To derive the claimed result, we derive a lower bound on the regret for the composite MDP presented above. Our analysis is largely built on the techniques used in the proof of \citep[Theorem~5]{jaksch2010near}. We also closely follow the notations used in \citep{jaksch2010near}.

Let us assume, as in the proof of \citep[Theorem~5]{jaksch2010near}, that all states $s_0$ are identified so that $M$ is equivalent to an MDP $M'$ with $kA'$ actions (note that following the same argument as in \citep{jaksch2010near}, despite the same maximal average reward, learning in $M'$ is easier than in $M$, and so any regret lower bound for $M'$ implies a lower bound in $M$, too). Note that by construction of $M$, it holds that $\mathbf V_{\max}$ in $M$ equals $\mathbf V_{\max}$ in $M'$. Denote by $(s^\star_0,a^\star)$ the \emph{good copy}, i.e., the one containing the good action $a^\star$. We assume that $a^\star$ is chosen uniformly at random among all actions $\{1,\ldots,k\}\times \{1,\ldots,A'\}$. Let $\EE_\star[\cdot]$ and $\EE_{\mathrm{unif}}[\cdot]$ respectively denote the expectation with respect to the random choice of $(s_0^\star,a^\star)$ and the expectation when there is no good action. Furthermore, let $\EE_a[\cdot]$ denote the expectation conditioned on $a=a^\star$, and introduce $N_1$, $N_0$, and $N_0^\star$ as the respective number of visits to $s_1$, $s_0$, and $(s_0,a^\star)$.

The proof proceeds in the same steps as in the proof of \citep[Theorem~5]{jaksch2010near} up to Equation (36) there, where it is shown that assuming that the initial state is $s_0$,
\begin{align*}
\EE_a[N_1] &\leq \EE_a[N_0 - N_0^\star] + (\delta + \varepsilon) D' \EE_a[N_0^\star] \leq T - \EE_{\mathrm{unif}}[N_1] + \varepsilon D' \EE_a [N_0^\star] \, ,\\
\EE_{\mathrm{unif}}[N_1] &\geq  \frac{T-D'}{2}\, ,
\end{align*}
so that the accumulated reward $R_{\bA,T}$ by the algorithm $\bA$ in $M'$ up to time step $T$ satisfies
\begin{align*}
\EE_a[R_{\bA,T}]  \leq  \EE_a[N_1] \leq \frac{T+D'}{2} + \varepsilon D' \EE_{a}[N_0^\star]\, .
\end{align*}

The following lemma, which is a straightforward modification to \citep[Lemma~13]{jaksch2010near}, enables us to control $\EE_a[N_0^\star]$:

\begin{lemma}
\label{lem:LB_Ea_Eunif}
Let $f : \{s_0,s_1\}^{T+1} \mapsto [0,B]$ be any function defined on any trajectory $\boldsymbol s_{T+1} = (s_t)_{1\leq t\leq T+1}$ in $M'$. Then, for any $\delta\in [0,\tfrac{1}{3}]$,  $\varepsilon \in (0, 1-2\delta)$, and $a\in \{1,\ldots,kA'\}$,
\begin{align*}
  \EE_a[f(\boldsymbol s)] &\leq \EE_{\mathrm{unif}}[f(\boldsymbol s)]  + \varepsilon B
  \sqrt{\frac{\log(2)\EE_{\mathrm{unif}}[N_0^\star]}{2(\delta + \varepsilon)(1-\delta-\varepsilon) }}\; .
\end{align*}
\end{lemma}

Noting that $N_0^\star$ is a function of $\boldsymbol s_{T+1}$ satisfying $N_0^\star \in [0,T]$, by Lemma \ref{lem:LB_Ea_Eunif} we deduce
\begin{align*}
\EE_a[N_0^\star] &\leq \EE_{\mathrm{unif}}[N_0^\star] + \varepsilon T
  \sqrt{\frac{\log(2)\EE_{\mathrm{unif}}[N_0^\star]}{2(\delta + \varepsilon)(1-\delta-\varepsilon) }}\\
&= \EE_{\mathrm{unif}}[N_0^\star] +   \varepsilon \Spanstar T \sqrt{\frac{\log(2)\EE_{\mathrm{unif}}[N_0^\star]}{2\mathbf V_{\max}}}
\, ,
\end{align*}
where we used  $\sqrt{\mathbf V_{\max}} = \Spanstar \sqrt{(\delta+\varepsilon)(1-\varepsilon-\delta)}$. As shown in the proof of \citep[Theorem~5]{jaksch2010near}, $\sum_{a=1}^{kA'} \EE_{\mathrm{unif}}[N_0^\star] \leq (T+D')/2$ and $\sum_{a=1}^{kA'} \sqrt{\EE_{\mathrm{unif}}[N_0^\star]} \leq \sqrt{kA'(T+D')/2}$, so that we finally get, using the relation $\EE_\star[R_{\bA,T}] = \frac{1}{kA'} \sum_{a=1}^{kA'} \EE_a[R_{\bA,T}]$,
\begin{align*}
\EE_\star[\kR_{\bA, T,M'}] &= \frac{\delta + \varepsilon}{2\delta+\varepsilon} T - \EE_\star[R_{\bA,T}] \\
&\geq \frac{\delta + \varepsilon}{2\delta+\varepsilon} T - \frac{T}{2} -\frac{\varepsilon D'T}{2 kA'} - \frac{\varepsilon D'^2}{2kA'} \\
&- \frac{\varepsilon^2\Spanstar D'T}{kA'}\sqrt{\frac{\log(2)kA'T}{4\mathbf V_{\max}}} - \frac{\varepsilon^2\Spanstar D'T}{ kA'}\sqrt{\frac{\log(2)kA'D'}{4\mathbf V_{\max}}} - \frac{D'}{2} \\
&\geq \frac{\varepsilon T}{4\delta + 2\varepsilon} -\frac{\varepsilon D'}{2 kA'}(T+D') -\frac{0.42\varepsilon^2 \Spanstar D'T  }{\sqrt{ kA'\mathbf V_{\max}}}(\sqrt{T} + \sqrt{D'} ) - \frac{D'}{2}\; .
\end{align*}

Noting that the assumption $T\geq  DSA$ implies $T\geq  16D'kA'$, we deduce that
\begin{align*}
\EE_\star[\kR_{\bA, T,M'}] &\geq
 \frac{\varepsilon T}{4\delta + 2\varepsilon} -\frac{\varepsilon D'T}{2 kA'}\Bigl(1 + \frac{1}{16kA'}\Big)
 -\frac{0.42\varepsilon^2 \Spanstar D'T\sqrt{T}  }{\sqrt{ kA'\mathbf V_{\max}}}\Bigl( 1 + \frac{1}{4\sqrt{kA'}} \Big) - \frac{D'}{2}\, .
\end{align*}

The first term in the right-hand side of the above satisfies
\begin{align*}
\frac{\varepsilon T}{4\delta + 2\varepsilon} &= \frac{\varepsilon T \Spanstar}{2} \geq  \frac{5\varepsilon \mathbf V_{\max} T}{6}\, ,
\end{align*}
since
\begin{align*}
\frac{\Spanstar}{\mathbf V_{\max}} = \frac{2\delta + \varepsilon}{ (\delta + \varepsilon )(1-\delta-\varepsilon)} \geq  1 + \frac{\delta}{\delta + \varepsilon} >\frac{5}{3}\, ,
\end{align*}
where we used $\varepsilon\leq \frac{\delta}{2}$ in the last step. Hence, we get
\begin{align*}
\EE_\star[\kR_{\bA, T,M'}] \geq  \frac{5}{6}\varepsilon \mathbf V_{\max} T -\frac{\varepsilon D'T}{2 kA'} \Bigl(1+\frac{1}{16kA'}\Big)- \frac{0.42\varepsilon^2 \Spanstar D'T\sqrt{T}  }{\sqrt{kA'\mathbf V_{\max}}}\Bigl(1 + \frac{1}{4\sqrt{kA'}} \Big) - \frac{D'}{2}\, .
\end{align*}
In particular, setting $\varepsilon = c\sqrt{\frac{kA'}{\mathbf V_{\max} T}}$ for some $c$ (which will be determined later) yields
\begin{align*}
\EE_\star[\kR_{\bA, T,M'}] &\geq  \frac{5}{6}c\sqrt{kA' \mathbf V_{\max} T} - \sqrt{kA' \mathbf V_{\max} T} \left(\frac{cD'}{2kA'\mathbf V_{\max}} \Bigl(1+\frac{1}{16kA'}\Big) \right) \\
&- \sqrt{kA' \mathbf V_{\max} T}\left( \frac{0.42c^2}{kA'} \frac{ D'\Spanstar}{\mathbf V_{\max}^2}\Bigl(1+\frac{1}{4\sqrt{kA'}}\Big)\right) - \frac{D'}{2}\, .
\end{align*}

To simplify the above bound, note that
\begin{align}
\label{eq:Dprime_over_Vmax}
\frac{D'}{\mathbf V_{\max}} &\leq \frac{(2\delta + \varepsilon)^2}{\delta(\delta+\varepsilon)(1-\delta-\varepsilon)} \leq 2\Bigl(\frac{2\delta + \epsilon}{\delta}\Big)^2 \leq 12.5\, ,
\end{align}
where we used $1-\varepsilon - \delta \geq  \tfrac{1}{2}$ since $\varepsilon\leq \frac{\delta}{2}$. Moreover,
\begin{align*}
\frac{D'\Spanstar}{\mathbf V_{\max}^2} &= \frac{D'\Spanstar}{\Spanstar^4 (\delta+\varepsilon)^2(1-\delta - \varepsilon)^2}  \\
&= \frac{(2\delta + \varepsilon)^3}{\delta (\delta+\varepsilon)^2(1-\delta - \varepsilon)^2}  \leq 4\Bigl(\frac{2\delta + \varepsilon}{\delta}\Big)^3 \leq 62.5\, .
\end{align*}
Putting these together with the fact that
\begin{align*}
\frac{D'}{2}\leq \frac{\sqrt{D'}}{2} \sqrt{\frac{T}{16kA'}} \leq \frac{\sqrt{12.5/16}}{2}\sqrt{\frac{\mathbf V_{\max}T}{kA'}} \leq 0.45\sqrt{\frac{\mathbf V_{\max}T}{kA'}}\, ,
\end{align*}
which follows from (\ref{eq:Dprime_over_Vmax}),
we deduce that
\begin{align*}
\EE_\star[\kR_{\bA, T,M'}] \geq  \sqrt{kA' \mathbf V_{\max} T}\Bigl(\frac{5c}{6} - \frac{12.5c}{2kA'} - \frac{12.5c}{32(kA')^2}
- \frac{26.25 c^2}{kA'} - \frac{6.6c^2}{(kA')^{3/2}} - \frac{0.45}{kA'}\Big),
\end{align*}
Taking $c=0.132$ and using the facts $k=\lfloor \tfrac{S}{2}\rfloor \geq  5$ and $A'=\lfloor\frac{A-1}{2}\rfloor \geq  4$ yield the announced result. This completes the proof provided that we show that this choice of $c$ satisfies $\varepsilon\leq \frac{\delta}{2}$. To this end, observe that by the assumption $T\geq  DSA \geq  \frac{16kA'}{\delta}$, it follows that
\begin{align*}
  \epsilon = 0.132\sqrt{\frac{kA'}{\mathbf V_{\max} T}} \leq \frac{0.132}{4}\sqrt{\frac{\delta} {\mathbf V_{\max} }} \leq \frac{0.132}{4}\sqrt{\frac{\delta(2\delta+\epsilon)^2}{(\delta+\epsilon)(1-\delta - \epsilon) }} \leq 0.047 (2\delta+\epsilon)\, ,
\end{align*}
so that $\epsilon \leq 0.1 \delta$. This concludes the proof.
\ep

\subsection{Proof of Lemma \ref{lem:LB_Ea_Eunif}}

The lemma follows by a slight modification of the proof of \citep[Lemma~13]{jaksch2010near}. We recall that according to Equations (49)-(51) in \citep{jaksch2010near},
\begin{align}
\label{eq:tmp_lemma_LB_Ea}
\EE_a[f(\boldsymbol s)] - \EE_{\mathrm{unif}}[f(\boldsymbol s)] &\leq \frac{B}{2} \sqrt{2 \log(2) \KL(\mathbb P_{\mathrm{unif}}, \mathbb \PP_a)}\,  ,
\end{align}
where
\begin{align*}
\KL(\mathbb P_{\mathrm{unif}}, \mathbb \PP_a) &= \sum_{t=1}^T \KL(\mathbb P_{\mathrm{unif}}(s_{t+1}|\boldsymbol s^t), \PP_a(s_{t+1}|\boldsymbol s^t)) \\
&= \sum_{t=1}^T \PP_{\mathrm{unif}}(s_t = s_0, a_t = a) \left(\delta \log\Bigl(\frac{\delta}{\delta + \varepsilon}\Big) + (1-\delta) \log\Bigl(\frac{1-\delta}{1-\delta - \varepsilon}\Big)\right)\, .
\end{align*}
Now using the inequality $\klber(a,b) \leq \frac{(a-b)^2}{b(1-b)}$ valid for all $a,b\in (0,1)$ (instead of \citep[Lemma~20]{jaksch2010near}) and noting that $\EE_{\mathrm{unif}}[N_0^\star]=\sum_{t=1}^T \PP_{\mathrm{unif}}(s_t = s_0, a_t = a)$, we obtain
\begin{align*}
\KL(\mathbb P_{\mathrm{unif}}, \mathbb \PP_a)= \klber(\delta , \delta+\varepsilon) \EE_{\mathrm{unif}}[N_0^\star]
\leq \frac{\varepsilon^2}{(1-\delta)(1-\delta-\varepsilon)} \EE_{\mathrm{unif}}[N_0^\star]\, .
\end{align*}
Plugging this into (\ref{eq:tmp_lemma_LB_Ea}) completes the proof.
\ep

\section{Concentration Inequalities}
\label{app:concentration}

\subsection{Proof of Lemma~\ref{lem:transportation}}
	Let us recall the fundamental equality
	\beqan
	\forall \lambda \in\Real,\quad
	\log \Esp_{P}[\exp(\lambda (X - \Esp_P[X])] = \sup_{Q \ll P} \bigg[ \lambda\Big(\Esp_{Q}[X]-\Esp_P[X]\Big) - \KL(Q,P) \bigg].
	\eeqan
	In particular, we obtain on the one hand that (see also \citep[Lemma~2.4]{boucheron2013concentration})
	\beqan
	\forall Q \ll P,\quad
	\Esp_{Q}[f]-\Esp_P[f] &\leq & \min_{\lambda \in \Real^+}  \frac{\phi_f(\lambda) + \KL(Q,P)}{\lambda}\, .
	\eeqan
	Since $\phi_f(0)=0$, then the right-hand side of the above is non-negative. Let us call it $u$.
	Now, we note that for any $t$ such that $u\geq t\geq 0$, by construction of $u$, it holds that $\KL(Q,P) \geq \phi_{\star,f}(t)$. Thus, $\{ x \geq 0 : \phi_{f,\star}(x) > \KL(Q,P) \} = (u,\infty)$
	and hence, $u = \phi_{+,f}^{-1}(\KL(Q,P))$.
	
	On the other hand, it holds
	\beqan
	\forall Q \ll P,\quad
	\Esp_{Q}[f]-\Esp_P[f] &\geq & \max_{\lambda \in \Real^-}  \frac{\phi_f(\lambda) + \KL(Q,P)}{\lambda}\, .
	\eeqan	
	Since $\phi(0)=0$, then the right-hand side quantity is non-positive. Let us call it $v$.
	Now, we note that for any $t$ such that $v\leq t\leq 0$, by construction of $v$, it holds that $\KL(Q,P) \geq \phi_{\star,f}(t)$. Thus, $\{ x \leq 0 : \phi_{\star,f}(x) > \KL(Q,P) \} = (-\infty,v)$
	and hence, $v = \phi_{-,f}^{-1}(\KL(Q,P))$.	
\ep

\subsection{Proof of Corollary~\ref{cor:transportationII}}
By a standard Bernstein argument (see for instance \citep[Section~2.8]{boucheron2013concentration}), it holds	
	\beqan
	\forall \lambda\in[0,3/\bS(f)),\quad
	\phi_f(\lambda)  &\leq& \frac{\Var_P[f]}{2}\frac{\lambda^2}{1- \frac{\bS(f) \lambda}{3}}\, , \\
	\forall x\geq 0,\quad	\phi_{\star,f}(x)
	&\geq& \frac{x^2}{2(\Var_P[f] + \frac{\bS(f)}{3} x )}\, .
	\eeqan
	Then, a direct computation (solving for $x$ in $\phi_{\star,f}(x)=t$) shows that
	\beqan
	\phi_{+,f}^{-1}(t) &\leq& \frac{\bS(f)}{3} t + \sqrt{ 2t\Var_P[f]  + \Big(\frac{\bS(f)}{3} t\Big)^2}
	\leq\sqrt{ 2t\Var_P[f]} +\frac{2}{3} t\bS(f) \,  ,\\
	\phi_{-,f}^{-1}(t)& \geq& \frac{\bS(f)}{3} t - \sqrt{ 2t\Var_P[f]  + \Big(\frac{\bS(f)}{3} t\Big)^2} \geq -\sqrt{ 2t\Var_P[f]}\,,
	\eeqan	
	where we used that $\sqrt{a+b} \leq \sqrt{a}+\sqrt{b}$ for $a,b\geq 0$.
	Combining these bounds,  we get
	\beqan
	\Esp_{Q}[f]-\Esp_P[f] &\leq&\sqrt{ 2\Var_P[f]\KL(Q,P) }+\frac{2}{3} \bS(f)\KL(Q,P)\, , \\
	\Esp_{P}[f]-\Esp_Q[f]	&\leq& \sqrt{ 2\Var_P[f]\KL(Q,P)}\,.
	\eeqan
\ep

\subsection{Proof of Lemma \ref{lem:KL_var}}
If $\EE_Q[f] \leq \EE_P[f]$, then the result holds trivially. We thus assume that $\EE_Q[f] > \EE_P[f]$. 
It is straightforward to verify that
\begin{align}
\EE_Q[f] - \EE_P[f] 
&= \sum_{x: Q(x)\geq P(x)} (f(x) - \EE_Q[f])(Q(x) - P(x))  + \sum_{x: Q(x)< P(x)} (f(x) - \EE_P[f])(Q(x) - P(x)) \sk
\label{eq:lem_pre_1}
&+ \sum_{x:P(x)>Q(x)} (\EE_P[f] - \EE_Q[f])(Q(x) - P(x)) \; .
\end{align}

The first term in the right-hand side of (\ref{eq:lem_pre_1}) is upper bounded as
\begin{align}
\sum_{x: Q(x) \geq  P(x) } (f(x) - \EE_Q[ f]) &(Q(x) - P(x) )= \sum_{x: Q(x) \geq   P(x) } \sqrt{Q(x)}(f(x) - \EE_Q[ f]) \frac{Q(x) - P(x) }{\sqrt{Q(x)}} \sk
&\stackrel{(a)}{\leq} \sqrt{\sum_{x: Q(x) \geq  P(x) } Q(x)(f(x) - \EE_Q[ f])^2}\sqrt{\sum_{x: Q(x) \geq  P(x) } \frac{(Q(x) - P(x) )^2}{Q(x)}} \sk
\label{eq:KL_var_1}
&\stackrel{(b)}{\leq} \sqrt{\Vcal_{Q,P}( f)}\sqrt{2\kl(P, Q)} \; ,
\end{align}
where (a) follows from Cauchy-Schwarz inequality and (b) follows from Lemma \ref{lem:refined_Pinsker_vector}.

Similarly, the second term in (\ref{eq:lem_pre_1}) satisfies
\begin{align}
\sum_{x: Q(x)< P(x) } (f(x) - \EE_P[ f])(Q(x) - P(x) )&= \sum_{x: Q(x) <  P(x) } \sqrt{P(x)}(f(x) - \EE_P[ f]) \frac{Q(x) - P(x) }{\sqrt{P(x)}} \sk
\label{eq:KL_var_2}
&\leq \sqrt{\Vcal_{P,Q}( f)}\sqrt{2\kl(P, Q)} \; .
\end{align}

Finally, we bound the last term in (\ref{eq:lem_pre_1}):
\begin{align}
(\EE_P[f] - \EE_Q[f])\sum_{x:P(x)>Q(x)}(Q(x) - P(x)) &\stackrel{(a)}{=} \frac{1}{2}(\EE_Q[f] - \EE_P[f])\|P-Q\|_1 \sk
\label{eq:KL_var_3}
&\leq  \frac{1}{2}\Span(f) \|P-Q\|_1^2 \stackrel{(b)}{\leq} \Span(f)\kl(P,Q) \;,
\end{align}
where (a) follows from the fact that for any pair of distributions $U, V\in \Pcal(\cX)$, it holds that $\sum_{x\in \Xcal} |U(x) - V(x)| = 2\sum_{x:U(x)\geq  V(x)} (U(x) - V(x))$, and where (b) follows from Pinsker's inequality.
The proof is concluded by combining (\ref{eq:KL_var_1}), (\ref{eq:KL_var_2}), and (\ref{eq:KL_var_3}).
\ep

\subsection{Proof of Lemma \ref{lem:Vcal_properties}}
Statement (i) is a direct consequence of the definition of $\Vcal_{P,Q}$. We next prove statement (ii). Observe that Lemma \ref{lem:refined_Pinsker_vector} implies that for all $x\in \Xcal$
\begin{align*}
|P(x) - Q(x)| \leq \sqrt{2\max(P(x), Q(x)) \kl(Q, P)} \; .
\end{align*}
Hence,
\begin{align}
\Vcal_{P,Q}(f) &= \sum_{x: P(x)\geq  Q(x)} P(x)(f (x) - \EE_{P}[f])^2 \sk
\label{eq:Vcal_1}
&\leq  \sum_{x:P(x)\geq  Q(x)} Q(x)(f(x) - \EE_{P}[f])^2 + \sqrt{2\kl(Q,P)}\sum_{x: P(x)\geq  Q(x)} \sqrt{P(x)}(f(x) - \EE_{P}[ f])^2 \; .
\end{align}

The first term in the right-hand side of (\ref{eq:Vcal_1}) is bounded as follows:
\begin{align*}
\sum_{x:P(x)\geq  Q(x)} Q(x)(f(x) - \EE_{P}[f])^2 &\leq 2\sum_{x:P(x)\geq  Q(x)} Q(x)(f (x) - \EE_{Q}[f])^2  + 2 (\EE_Q[f] - \EE_P[f])^2 \sk
&\leq 2\Var_{Q}(f) + 2 (\EE_Q[f] - \EE_{P}[f])^2 \; .
\end{align*}
Note that
$$
(\EE_{Q}[f] - \EE_P[f])^2 \leq \Span(f)^2 \|P - Q\|_1^2 \leq 2\Span(f)^2\kl(Q,P)\; ,
$$
which further gives
\begin{align*}
\sum_{x:P(x)\geq  Q(x)} Q(x)(f (x) - \EE_{P}[f])^2 &\leq
2\Var_{Q}(f) + 4\Span(f)^2\kl(Q,P)  \; .
\end{align*}

Now we consider the second term in (\ref{eq:Vcal_1}). First observe that
\begin{align*}
\sum_{x:  P(x)\geq  Q(x)} \sqrt{P(x)}(f(x) - \EE_{P}[ f])^2 &\leq \sqrt{\sum_{x: P(x)\geq  Q(x)} P(x)(f(x) - \EE_{P}[ f])^2} \sqrt{\sum_{x}(f(x) - \EE_{P}[ f])^2} \\
&\leq \sqrt{\Vcal_{P,Q}(f)}\Span(f)\sqrt{|\Xcal|} \; ,
\end{align*}
thanks to Cauchy-Schwarz inequality. Hence, the second term in (\ref{eq:Vcal_1}) is upper bounded by $$\Span(f)\sqrt{2|\Xcal|\Vcal_{P,Q}(f) \kl(Q,P)}\; .$$

Combining the previous bounds together, we get
\begin{align*}
\Vcal_{P,Q}(f) \leq 2\Var_{Q}(f) + 4\Span(f)^2\kl(Q,P) + \Span(f)\sqrt{2|\Xcal|\Vcal_{P,Q}(f)\kl(Q, P)} \; ,
\end{align*}
which leads to
\begin{align*}
\Bigl(\sqrt{\Vcal_{P,Q}(f)} - \Span(f)\sqrt{|\Xcal|\kl(Q, P)/2}\Big)^2 \leq 2\Var_{Q}(f) + \Span(f)^2(|\Xcal|/2+4)\kl(Q,P) \; ,
\end{align*}
so that using the inequality $\sqrt{a+b}\leq \sqrt{a}+\sqrt{b}$, we finally obtain
\begin{align*}
\sqrt{\Vcal_{Q,P}(f)} &\leq \sqrt{2\Var_{Q}(f) + \Span(f)^2(|\Xcal|/2+4)\kl(Q,P)} + \Span(f)\sqrt{|\Xcal|\kl(Q, P)/2}\\
&\leq \sqrt{2\Var_{Q}(f)} + \Span(f)(\sqrt{2|\Xcal|}+2)\sqrt{\kl(Q,P)}
 \; .
\end{align*}
The proof is completed by observing that $\sqrt{2|\Xcal|}+2 \leq 3\sqrt{|\Xcal|}$ for $|\Xcal|\geq  2$.
\ep

\subsection{Proof of Lemma \ref{lem:Var_phat}}
Let $\delta \in (0,1)$ and $(s,a) \in \Scal\times \Acal$. Consider an episode $k\geq 1$ such that $M\in \Mcal_k$, and define $\hat p_k = \hat p_k(\cdot|s,a)$, $p = p(\cdot|s,a)$, and $N_k = N_k(s,a)$.
Observe that by a Bernstein-like inequality \citep[Lemma~F.2]{dann2017unifying}, we have: for all $s'\in \Scal$, with probability at least $1-\delta$,
\begin{align*}
\hat p_k(s') - p(s') \le \sqrt{\frac{2p(s')C_b}{N_k}} + \frac{2C_b}{N_k} \; ,
\end{align*}
with $C_b = C_b(t,\delta) := \log(3\log(\max(e,t))/\delta)$. It then follows that with probability at least $1-\delta$,
\begin{align}
\Var_{\hat p_k}&(f)= \sum_{s'}\hat p_k(s')(f(s') - \EE_{\hat p_k}[f])^2 \sk
&\le \sum_{s'}p(s')(f(s') - \EE_{\hat p_k}[f])^2 + \sqrt{\frac{2C_b}{N_k}}\sum_{s'}\sqrt{p(s')}(f(s') - \EE_{\hat p_k}[f])^2 + \frac{2C_b}{N_k}\sum_{s'}(f(s') - \EE_{\hat p_k}[f])^2 \sk
\label{eq:Var_phat_1}
&\le \underbrace{\sum_{s'}p(s')(f(s') - \EE_{\hat p_k}[f])^2}_{Z_1} + \sqrt{\frac{2C_b}{N_k}}\underbrace{\sum_{s'}\sqrt{p(s')}(f(s') - \EE_{\hat p_k}[f])^2}_{Z_2}
+\frac{2C_b S\Span(f)^2}{N_k}
\; .
\end{align}

Next we bound $Z_1$ and $Z_2$. Observe that
\begin{align*}
Z_1 &\le 2\sum_{s'}p(s')(f(s') - \EE_{p}[f])^2 + 2(\EE_p[f] - \EE_{\hat p_k}[f])^2\\
&\le 2\Var_p(f) + 4\Span(f)^2 \kl(\hat p_k, p)\; ,
\end{align*}
where the last inequality follows from
\begin{align}
\label{eq:Var_phat_2}
(\EE_p[f] - \EE_{\hat p_k}[f])^2 \le \Span(f)^2 \|p - \hat p_k\|_1^2 \le 2\Span(f)^2 \kl(\hat p_k, p)\; .
\end{align}

For $Z_2$ we have
\begin{align*}
Z_2 \le 2\sum_{s'}\sqrt{p(s')}(f(s') - \EE_p[f])^2 + 2 (\EE_p[f] - \EE_{\hat p_k}[f])^2 \sum_{s'} \sqrt{p(s')} \; .
\end{align*}
Now, using Cauchy-Schwarz inequality
\begin{align*}
\sum_{s'}\sqrt{p(s')}(f(s') - \EE_{p}[f])^2 &\le \sqrt{\sum_{s'}p(s')(f(s') - \EE_p[f])^2 \sum_{s'} (f(s') - \EE_p[f])^2 } \\
&\le \sqrt{S\Var_p(f)} \Span(f) \; ,
\end{align*}
so that using (\ref{eq:Var_phat_2}), we deduce that
\begin{align*}
Z_2 &\le 2\Span(f)\sqrt{S\Var_p(f)}  + 4\Span(f)^2\kl(\hat p_k,p)\sum_{s'} \sqrt{p(s')} \\
&\le 2\Span(f)\sqrt{S\Var_p(f)}  + 4\Span(f)^2\kl(\hat p_k,p)\sqrt{S}
\; ,
\end{align*}
where the last inequality follows from Jensen's inequality:
$$
\sum_{s'} \sqrt{p(s')} = \sum_{s'} p(s')\sqrt{\frac{1}{p(s')}} \le \sum_{s'} \sqrt{\frac{p(s')}{p(s')}} = \sqrt{S}\; .
$$

Putting together, we deduce that  with probability at least $1-\delta$,
\begin{align*}
\Var_{\hat p_k}(f) &\le 2\Var_p(f) + 2\Span(f)\sqrt{\frac{2SC_b}{N_k}} \Bigl(\sqrt{\Var_p(f)}  + 2\Span(f)\kl(\hat p_k, p)\Big) + \Span(f)^2\Bigl(4\kl(\hat p_k, p) +  \frac{2SC_b}{N_k}\Big)\; .
\end{align*}
Noting that $M\in \Mcal_k$, we obtain
\begin{align*}
\Var_{\hat p_k}(f) &\le 2\Var_p(f) +  \Span(f)\sqrt{\frac{8S\Var_p(f)C_b}{N_k}} + 4\Span(f)^2\frac{\sqrt{2SC_b}C_p}{N_k^{3/2}} + \frac{(4C_p+ 2SC_b)\Span(f)^2}{N_k} \\
&\le 2\Var_p(f) +  \Span(f)\sqrt{\frac{8S\Var_p(f)C_b}{N_k}} + \frac{S\Span(f)^2}{N_k}(16B\sqrt{2SC_b} + 16B+ 2C_b) \\
&\le 2\Var_p(f) +  \Span(f)\sqrt{\frac{8S\Var_p(f)B}{N_k}} + \frac{36S^{3/2}B^{3/2}\Span(f)^2}{N_k}
\; ,
\end{align*}
with probability at least $1-\delta$, where we used $C_p = 4SB$, $C_b\le B$, and $S\geq 2$. The proof is concluded by observing that
\begin{align*}
\sqrt{\Var_{\hat p_k}(f)} &\le \sqrt{2\Var_{p}(f)} +  \Span(f)\sqrt{\frac{SB}{N_k}} + 6\Span(f)B\sqrt{\frac{S^{3/2}}{N_k}} \\
&\le \sqrt{2\Var_{p}(f)} +  \frac{6S\Span(f)B}{\sqrt{N_k}}
\; ,
\end{align*}
with probability at least $1-\delta$.
\ep

\section{Regret Upper Bound for \KLUCRL}
\label{app:Reg_UB}
In this section, we provide the proof of the main result
(Theorem \ref{thm:klucrl_reg}). We will try to closely follow the notations used in the proof of \cite[Theorem~2]{jaksch2010near}.

We first recall the following result indicating that the true model belongs to the set of plausible MDPs with high probability. Recall that for $\delta\in (0,1]$ and $t\in \NN$,
\begin{align*}
C_\mu:= C_\mu(T,\delta) &= \log(4SA\log(T)/\delta)/1.99\, , \\
C_p:=C_p(T,\delta) &= S\left(B+\log(G)(1+1/G)\right)\, ,
\end{align*}
where
\begin{align}
\label{eq:B_T_delta}
B:&=B(T,\delta)=\log(2eS^2A\log(T)/\delta)\, ,\\
G:&=G(T,\delta)=B+1/\log(T)\, . \nonumber
\end{align}
Moreover, observe that $C_p\leq 4SB$.

\begin{lemma}[{{\citep[Proposition~1]{filippi2010optimism}}}]
\label{lem:concentration}
For all $T\geq  1$ and $\delta>0$, and for any pair $(s,a)$, it holds that
\begin{align*}
\PP\Bigl(\forall t\leq T,\; |\hat \mu_t(s,a)-\mu(s,a)| \leq \sqrt{C_\mu/N_t(s,a)}\Big) &\geq  1-\frac{\delta}{SA}\, ,\\
\PP\Bigl(\forall t\leq T,\; N_t(s,a)\emph{\kl}(\hat p_t(s,a),p(\cdot|s,a)) \leq C_p \Big) &\geq  1-\frac{\delta}{SA}\,  .
\end{align*}
In particular, $\PP(\forall t\leq T, \; M\in \Mcal_t)\geq  1-2\delta$.
\end{lemma}

Next we prove the theorem.\\

\noindent\emph{Proof (of Theorem \ref{thm:klucrl_reg}).}
Let $T\geq  1$ and $\delta\in (0,1)$. Fix algorithm $\bA = \KLUCRL$. Denote by $m(T)$ the number of episodes started by \KLUCRL\ up to time step $T$ (hence, $1\leq k\leq m(T)$). 

By applying Azuma-Hoeffding inequality, as in the proof of \citep[Theorem~2]{jaksch2010near}, we deduce that
\begin{align*}
\kR_{\bA,T} &= Tg^\star  - \sum_{t=1}^T r(s_t,a_t) \leq \sum_{s,a} N_T(s,a)(g^\star - \mu(s,a)) + \sqrt{\tfrac{1}{2}T\log(1/\delta)} \; ,
\end{align*}
with probability at least $1-\delta$.
The regret up to time $T$ can be decomposed as the sum of the regret incurred in various episodes. Let $\Delta_k$ denote the regret in episode $k$:
\begin{align*}
\Delta_k:=\sum_{s,a} v_k(s,a)(g^\star - \mu(s,a))\; .
\end{align*}
Therefore, Lemma \ref{lem:concentration} implies that with probability at least $1-3\delta$,
\begin{align*}
\kR_{\bA,T} &\leq \sum_{k=1}^{m(T)} \Delta_k \bI \{M\in \Mcal_k\} + \sqrt{\tfrac{1}{2}T \log(1/\delta)} \; .
\end{align*}

Next we derive an upper bound on the first term in the right-hand side of the above inequality. Consider an episode $k\geq 1$ such that $M\in \Mcal_k$.
The state-action pair $(s,a)$ is considered as \emph{sufficiently sampled} in episode $k$ if its number of observations satisfies $N_k(s,a)\geq  \ell_{s,a}$, with
$$
\ell_{s,a}=\ell_{s,a}(T,\delta):= \max\Bigl\{\frac{128SB \max(\Psi^2,1)}{\phi(s,a)^2},\, 32SB\Big(\frac{\log(D)}{\log(1/\gamma)}\Big)^2\Big\}, \quad \forall s,a \, ,
$$
where  $B$ is given in (\ref{eq:B_T_delta}), and where $\gamma$ denotes the contraction factor of the mapping induced by the transition probability matrix $P_\star$ of the optimal policy ($\gamma$ can be determined as a function of elements of $P_\star$).

Now consider the case where all state-action pairs are sufficiently sampled in episode $k$ (we analyse the case where some pairs are under-sampled (i.e., not sufficiently sampled) at the end of the proof). We have
\begin{align*}
|\tilde \mu_k(s,a) - \mu(s,a)| &\le |\tilde \mu_k(s,a) - \hat \mu_k(s,a)| + |\hat \mu_k(s,a) - \mu(s,a)|
\le 2\sqrt{\frac{C_\mu}{N_k(s,a)^+}}\; .
\end{align*}

Hence,
\begin{align*}
\Delta_k &= \sum_{s,a} v_k(s,a)(g^\star - \tilde \mu_k(s,a)) + \sum_{s,a} v_k(s,a)(\tilde \mu_k(s,a) - \mu(s,a)) \\ 
&\leq \sum_{s,a} v_k(s,a)(g^\star - \tilde \mu_k(s,a)) + 2\sqrt{ C_\mu} \sum_{s,a}\frac{v_k(s,a)}{\sqrt{N_k(s,a)^+}}\; .
\end{align*}

Let $\tilde \mu_k$ and $\widetilde {P}_k$ respectively denote the reward vector and transition probability matrix induced by  the policy $\tilde \pi_k$ on $\tilde M_k$, i.e., $\tilde \mu_k := (\tilde \mu_k(s,\tilde\pi_k(s)))_s$, $\widetilde {P}_k:= \bigl(\tilde p_k(s'|s,\tilde\pi_k(s))\big)_{s,s'}$. By Bellman optimality equation, $\tilde g_k - \tilde \mu_k(s,a) = (\widetilde {P}_k-\bsym I) \tilde b_k$. Hence, defining $v_k = (v_k(s,\tilde\pi_k(s))_s$ yields
\begin{align*}
\Delta_k \leq v_k(\widetilde {P}_k - I )\tilde b_k + (g^\star - \tilde g_k)v_k\mathbf 1 + 2\sqrt{C_\mu} \sum_{s,a}\frac{v_k(s,a)}{\sqrt{N_k(s,a)^+}}\; .
\end{align*}


Now we use the following decomposition:
\begin{align*}
v_k( \widetilde {P}_k -I) \tilde b_k &=  \underbrace{v_k(\widetilde{P}_k - {P}_k) b^\star}_{F_1(k)}
+ \underbrace{v_k(\widetilde {P}_k - {P}_k) (\tilde b_k - b^\star)}_{F_2(k)}
+ \underbrace{v_k({P}_k - {I}) \tilde b_k }_{F_3(k)}\; .
\end{align*}


Let $c=1+\sqrt{2}$. The following two lemmas provide upper bounds for $F_1(k)$ and $F_2(k)$:

\begin{lemma}
\label{lem:F1}
For all $k\in \NN$ such that $M\in \Mcal_k$, with probability at least $1-\delta$, it holds that
\begin{align*}
F_1(k) &\leq (4+6\sqrt{2})\sqrt{SB}\sum_{s,a} v_k(s,a) \sqrt{\frac{\mathbf V^\star_{s,a}}{N_k(s,a)^+}} + 63\Spanstar S^{3/2} B^{3/2}\sum_{s,a}\frac{v_k(s,a)}{N_k(s,a)^+}\; .
\end{align*}
\end{lemma}


\begin{lemma}
\label{lem:F2}
Let $k\in \NN$ be the index of an episode such that $M\in \Mcal_k$. Assuming that $N_k(s,a)\geq \ell_{s,a}$ for all $s,a$, it holds that
\begin{align*}
F_2(k) + (g^\star - \tilde g_k)v_k\mathbf 1 &\le
\bigl(2 \sqrt{32SB} +1\big) \sum_{s,a} \frac{v_k(s,a)}{\sqrt{N_k(s,a)^+}}\, .
\end{align*}
\end{lemma}

\paragraph{Analysis of Term $F_3$.}Now we bound the term $\sum_{k=1}^{m(T)} F_3(k)$. To this end, similarly to the proof of \citep[Theorem~2]{jaksch2010near} and
\citep[Theorem~1]{filippi2010optimism}, we define the martingale difference sequence $(Z_t)_{t\geq 1}$, where $Z_t=(p(\cdot|s_t,a_t) - \mathbf e_{s_{t+1}})\tilde b_{k(t)} \bI\{M\in  \cM_{k(t)}\}$ for $t\in \{t_k, t_{k+1}-1\}$, where $k(t)$ denotes the episode containing $t$.
Note that for all $t$, $|Z_t| \leq 2D$.
Now applying  Azuma-Hoeffding inequality, we deduce that with probability at least $1-\delta$
\begin{align*}
\sum_{k=1}^{m(T)} F_3(k) &\leq \sum_{t=1}^T Z_t + 2m(T) D\\
&\leq
D\sqrt{2T\log(1/\delta)} + 2DSA \log_2\bigl(\tfrac{8T}{SA}\big) \;.
\end{align*}

\paragraph{The regret due to under-sampled state-action pairs.}To analyze the under-sampled regime, where some state-action pair is not sufficiently sampled, we borrow some techniques from \citep{auer2007logarithmic}. 
For any state-action pair $(s,a)$, let $L_{s,a}$ denote the set of indexes of episodes in which $(s,a)$ is chosen and yet $(s,a)$ is under-sampled; namely $k\in L_{s,a}$ if  $\tilde \pi_k(s) = a$ and $N_k(s,a)\leq \ell_{s,a}$. Furthermore, let $\tau_k(s,a)$ denote the length of such an episode.

Consider an episode $k\in L_{s,a}$.
By Markov's inequality, with probability at least $\frac{1}{2}$, it takes at most $2T_M$ to reach state $s$ from any state $s'$ in $k$, where $T_M$ is the mixing time of $M$. Let us divide episode $k$ into $\lfloor\frac{\tau_k(s,a)}{2T_M}\rfloor$ sub-episodes, each with length greater than $2T_M$. It then follows that in each sub-episode, $(s,a)$ is visited with probability at least $\frac{1}{2}$.

Using Hoeffding's inequality, if we consider $n$ such sub-episodes, with probability at least $1-\frac{\delta}{SA}$,
\begin{align*}
N(s,a) > n/2 - \sqrt{n\log(SA/\delta)}.
\end{align*}
Now we find $n$ that implies $N(s,a)<\ell_{s,a}$. Noting that $x\mapsto \frac{x}{2} - \sqrt{\alpha x}$ is increasing for $x\geq \alpha$, we have that for $n>10\max(\ell_{s,a},\log(SA/\delta))$,
\begin{align*}
n/2 - \sqrt{n\log(SA/\delta)} &> 5\max(\ell_{s,a},\log(SA/\delta)) - \sqrt{10\max(\ell_{s,a},\log(SA/\delta))\log(SA/\delta)}\\
&> \max(\ell_{s,a},\log(SA/\delta))\; .
\end{align*}
Hence, with probability at least $1-\frac{\delta}{SA}$, it holds that
\begin{align*}
\sum_{k\in L_{s,a}} \Bigl\lfloor \frac{\tau_k(s,a)}{2T_M} \Big\rfloor \leq 10 \max(\ell_{s,a},\log(SA/\delta))\, .
\end{align*}
Hence, the regret due to under-sampled state-action pairs can be upper bounded by
\begin{align*}
\sum_{s,a}\sum_{k\in L_{s,a}} \tau_k(s,a) &\leq 20T_M \sum_{s,a} \max(\ell_{s,a},\log(SA/\delta)) + 2T_M \sum_{s,a} |L_{s,a}| \\
&\leq 20T_M \sum_{s,a} \max(\ell_{s,a},\log(SA/\delta)) + 2T_MS^2A^2 \log_2\bigl(\tfrac{8T}{SA}\big)\; ,
\end{align*}
with probability at least $1-\delta$. Here we used that $|L_{s,a}|\le m(T)$.

Now applying Lemmas \ref{lem:F1} and \ref{lem:F2} together with the above bounds, and using the fact $C_\mu \le B/1.99$, we deduce that with probability at least $1-3\delta$
\begin{align*}
\sum_{k=1}^{m(T)} \Delta_k \bI \{M\in \Mcal_k\} &\le (4+6\sqrt{2})\sqrt{SB}\sum_{s,a} \frac{v_k(s,a)}{\sqrt{N_k(s,a)^+}} \sqrt{\mathbf V^\star_{s,a}} \\
&+ (2\sqrt{32SB}+3\sqrt{B} + 1)\sum_{s,a}
\frac{ v_k(s,a)}{\sqrt{N_k(s,a)^+}} \\
&+ 63\Spanstar S^{3/2} B^{3/2}\sum_{s,a}\frac{v_k(s,a)}{N_k(s,a)^+} \\
&+ D\sqrt{2T\log(1/\delta)} + 2DSA \log_2\bigl(\tfrac{8T}{SA}\big) \\ 
&+ 20T_M \sum_{s,a} \max(\ell_{s,a},\log(SA/\delta)) + 2T_MS^2A^2 \log_2\bigl(\tfrac{8T}{SA}\big) \; .
\end{align*}

To simplify the above bound, we will use Lemmas \ref{lem:seq_sqrt}, \ref{lem:seq_harmonic}, and \ref{lem:sum_square_prb} together with Jensen's inequality:
\begin{align*}
&\sum_{k=1}^{m(T)} \sum_{s,a}\frac{v_k(s,a)}{\sqrt{N_k(s,a)^+}}\le c\sum_{s,a}\sqrt{N_T(s,a)} \le c\sqrt{SAT}\; , \\
&\sum_{k=1}^{m(T)} \sum_{s,a}\frac{v_k(s,a)}{\sqrt{N_k(s,a)^+}}\sqrt{\mathbf V^\star_{s,a}}\le c\sum_{s,a}\sqrt{\mathbf V^\star_{s,a}N_T(s,a)} \le c\sqrt{T\textstyle\sum_{s,a} \mathbf V^\star_{s,a}}\; ,\\
&\sum_{k=1}^{m(T)} \sum_{s,a}\frac{v_k(s,a)}{N_k(s,a)^+}\le 2\sum_{s,a}\log(N_T(s,a)) + SA \le 2SA\log\bigl(\tfrac{T}{SA}\big) + SA \; .
\end{align*}

Putting everything together, we deduce that with probability at least $1-6\delta$,
\begin{align*}
\kR_{\bA,T}&\le \sum_{k=1}^{m(T)} \Delta_k \bI \{M\in \Mcal_k\} + \sqrt{\tfrac{1}{2}T\log(1/\delta)} \\
&\le 31\sqrt{S\sum_{s,a} \mathbf V^\star_{s,a} TB} + 35S\sqrt{ATB} + (\sqrt{2}D+1)\sqrt{T\log(1/\delta)} \\
&+ 126S^{5/2}A B^{5/2} \log\bigl(\tfrac{T}{SA}\big) + 2DSA \log_2\bigl(\tfrac{8T}{SA}\big) \\
&+ 20T_M \sum_{s,a} \max(\ell_{s,a},\log(SA/\delta)) + 2T_MS^2A^2 \log_2\bigl(\tfrac{8T}{SA}\big) + 63S^{5/2}A  \; .
\end{align*}
Hence,
\begin{align*}
\kR_{\bA,T}&\le 31\sqrt{S\sum_{s,a} \mathbf V^\star_{s,a} TB} + 35S\sqrt{ATB} + (\sqrt{2}D+1)\sqrt{T\log(1/\delta)} \\
&+ \widetilde \Ocal\Bigl(     SA(T_MSA + D + S^{3/2}) \log(T) \Big) \; .
\end{align*}
Noting that  $B=\Ocal(\log(\log(T)/\delta))$ gives the desired scaling and completes the proof.
\ep
\medskip

Next we prove Lemmas \ref{lem:F1} and \ref{lem:F2}.
\medskip

\subsection{Proof of Lemma \ref{lem:F1}}
We have
$$
F_1(k) = \underbrace{v_k(\widehat{P}_k - {P}_k) b^\star}_{G_1} + \underbrace{v_k(\widetilde{P}_k - \widehat {P}_k) b^\star}_{G_2}
$$

Next we provide upper bounds for $G_1$ and $G_2$.

\paragraph{Term $G_1$.}
We have
\begin{align*}
G_1 &=\sum_{s} v_k(s, \pi_k(s)) \sum_{s'} b^\star(s') \bigl(\hat p_k(s'|s, \pi_k(s)) - p(s'|s, \pi_k(s)) \big) \\
&\leq \sum_{s,a} v_k(s, a) \sum_{s'} b^\star(s') \bigl(\hat p_k(s'|s, a) - p(s'|s, a) \big)\; .
\end{align*}

Fix $s\in \Scal$ and $a\in \Acal$. Define the short-hands $p= p(\cdot|s,a)$, $\hat p_k=\hat p_k(\cdot|s,a)$, and $N_k^+ = N_k(s,a)^+$.
Applying Corollary \ref{cor:transportationII} (the first statement) and using the fact that $M\in \Mcal_k$ give:
\begin{align*}
\sum_{s'} b^\star(s') (\hat p_k(s') - p(s') ) &\leq \sqrt{2\mathbf V^\star_{s,a}\kl(\hat p_k, p)} + \frac{2}{3}\Spanstar \kl(\hat p_k,p) \\
&\leq \sqrt{8S\mathbf V^\star_{s,a}B/N_k^+} + \frac{8\Spanstar SB}{3N_k^+}
\;
.
\end{align*}
Therefore,
\begin{align*}
G_1 &\leq \sqrt{8SB}\sum_{s,a} v_k(s,a) \sqrt{\mathbf V^\star_{s,a}/N_k(s,a)^+} + \frac{8}{3}\Spanstar SB\sum_{s,a}v_k(s,a)/N_k(s,a)^+\; .
\end{align*}

\paragraph{Term $G_2$.}
We have
\begin{align*}
G_2 &\leq \sum_{s,a} v_k(s,a) \sum_{s'} b^\star(s') \bigl(\tilde p_k(s'|s,a) - \hat p_k(s'|s,a) \big)\;.
\end{align*}

Fix $s\in \Scal$ and $a\in \Acal$. Define the short-hands $\hat p_k= \hat p_k(\cdot|s,a)$, $\tilde p_k=\tilde p_k(\cdot|s,a)$, and $N_k^+ = N_k(s,a)^+$.
An application of Lemma \ref{lem:KL_var} and Lemma \ref{lem:Vcal_properties} gives
\begin{align*}
\sum_{s'} b^\star(s') (\tilde p_k(s') - \hat p_k(s') ) &\leq \Bigl(\sqrt{\Vcal_{\tilde p_k,\hat p_k}(b^\star)} + \sqrt{\Vcal_{\hat p_k,\tilde p_k}(b^\star)}\Big)\sqrt{2\kl(\hat p_k, \tilde p_k)} + \Spanstar \kl(\hat p_k,\tilde p_k) \\
&\leq c\sqrt{2\Var_{\hat p_k}(b^\star) \kl(\hat p_k, \tilde p_k)} + \Spanstar (1+ 3\sqrt{2S})\kl(\hat p_k, \tilde p_k)
\; ,
\end{align*}
where $c=1+\sqrt{2}$. Note that when $M\in \Mcal_k$, an application of Lemma \ref{lem:Var_phat} implies that, with probability at least $1-\delta$,
\begin{align*}
\sum_{s'} b^\star(s') (\tilde p_k(s') -\hat  p_k(s') ) &\leq 4c\sqrt{S\mathbf V^\star_{s,a} B/N_k^+}  + \frac{\Spanstar S^{3/2}B^{3/2}}{N_k^+}(12c\sqrt{2} + 12\sqrt{2} + 4/\sqrt{S}) \\
&\leq 4c\sqrt{S\mathbf V^\star_{s,a} B/N_k^+}  + \frac{61\Spanstar S^{3/2}B^{3/2}}{N_k^+}
\; ,
\end{align*}
where we used that $S\geq  2$. Multiplying by $v_k(s,a)$ and summing over $s,a$ yields
\begin{align*}
G_2 &\leq 4c\sqrt{SB}\sum_{s,a} v_k(s,a) \sqrt{\mathbf V^\star_{s,a}/N_k(s,a)^+} + 61\Spanstar S^{3/2} B^{3/2}\sum_{s,a}v_k(s,a)/N_k(s,a)^+\; .
\end{align*}
The lemma follows by combing bounds on $G_1$ and $G_2$.

%
\ep


\subsection{Proof of Lemma \ref{lem:F2}}
Let $k\geq 1$ be the index of an episode such that $M\in \Mcal_k$. Let $\tilde \star:=\tilde \star_k$ denote the optimal policy in $\cM_k$.
The proof proceeds in three steps.

\paragraph{Step 1.}
We remark that by definition of the bias functions, it holds that 
\begin{align*}
\tilde b_k - b^\star &= (g^\star - \tilde g_k)\mathbf 1 + \tilde \mu_k + \widetilde P_k b^\star - \mu_\star - P_\star b^\star + \widetilde P_k(\tilde b_k-b^\star)\\
&\le (\tilde g_{\tilde \star} - \tilde g_k)\mathbf 1 + \tilde \mu_k- \mu_k
+ (\widetilde P_k -P_k ) b^\star
+ \widetilde P_k(\tilde b_k-b^\star)  - \phi_k\,,
\end{align*}
where we define  $\phi_k(s):=\phi(s,\tilde \pi_k(s))$ for all $s$.
Defining
\begin{align*}
\xi_k(s) &= 2\sqrt{C_\mu/N_k(s,\tilde \pi_k(s))^+},\qquad
\zeta_k(s) =  \Psi\sqrt{32SB /N_k(s,\tilde \pi_k(s))^+} \;,
\end{align*}
we obtain the following bound:
\begin{align*}
\tilde b_k - b^\star  &\leq \frac{1}{\sqrt{t_k}}\mathbf 1 + \xi_k + \zeta_k- \phi_k + \widetilde P_k (\tilde b_k-b^\star)\,.
\end{align*}
It is straightforward to check that the assumption $N_k(s,\tilde \pi_k(s))\geq \ell_{s,\tilde \pi_k(s)}$ for all $s$  implies
\begin{align}
\label{eq:b_k_bstar_relation}
\tilde b_k - b^\star &\leq \widetilde P_k (\tilde b_k-b^\star)\,.
\end{align}
Note also that $\phi(s,\tilde \pi_k(s))\geq 0$ since $\star$ is $b^\star$-improving.

On the other hand, it holds that
\beqan
b^\star - 	\tilde b_{\tilde \star}&=& (\tilde g_{\tilde \star} - g^\star)\mathbf 1 +  \mu_\star + P_\star b^\star
- \tilde \mu_{\tilde\star} - \widetilde P_{\tilde \star} \tilde b_{\tilde \star}\\
&\leq& (\tilde g_{\tilde \star} - g^\star)\mathbf 1  +  \mu_\star + P_\star b^\star - \mu_\star - P_\star \tilde b_{\tilde \star}\\
&=&  (\tilde g_{\tilde \star} - g^\star)\mathbf 1 + P_\star (b^\star - \tilde b_{\tilde \star})\, .
\eeqan
Noting $P_\star \mathbf 1 = \mathbf 1$, and since all entries of $P_\star$ are non-negative, we thus get  for all $J\in \NN$,
\beqan
b^\star - 	\tilde b_{\tilde \star} &\leq& J(\tilde g_{\tilde \star} - g^\star)\mathbf 1  + P_\star^J(b^\star - \tilde b_{\tilde \star})\,.
\eeqan

\paragraph{Step 2.}
Let us now introduce $\cS_s^+ = \{ x \in \cS : \widetilde P_k (s,x) > P_k(s,x) \}$ as well as  its  complementary set $\cS_s^- =\cS\setminus\cS_s^+$. Using (\ref{eq:b_k_bstar_relation}), $\tilde b_k - b^\star\le 0$ so that
\begin{align*}
v_k(\widetilde P_k- P_k)(\tilde b_k - b^\star)
&= \sum_s v_k(s, \tilde \pi_k(s))\sum_{x\in\cS}
(\widetilde P_k (s,x) - P_k(s,x))(\tilde b_k(x) - b^\star(x))\\
&\leq \sum_s v_k(s,\tilde \pi_k(s))\sum_{x\in\cS_s^-}
(\underbrace{P_k (s,x) - \widetilde P_k(s,x)}_{\geq 0})(b^\star(x)-\tilde b_k(x) )\; .
\end{align*}

We thus obtain
\begin{align}
v_k(\widetilde P_{k}- P_{{k}})&(\tilde b_{k} - b^\star)
\leq \sum_s
v_k(s,\tilde \pi_k(s))\sum_{x\in\cS_s^-}
(P_{k} (s,x) - \widetilde P_{k}(s,x))(b^\star(x) - \tilde b_{\tilde \star}(x))\nonumber\\
&+
\sum_s
v_k(s,\tilde \pi_k(s))\sum_{x\in\cS_s^-}
(P_{k} (s,x) - \widetilde P_{k}(s,x))(\tilde b_{\tilde\star}(x) - \tilde b_{k}(x))\nonumber\\
&\le
\sum_s
v_k(s,\tilde \pi_k(s))\sum_{x\in\cS_s^-}
(P_{k} (s,x) - \widetilde P_{k}(s,x))[P_\star^J(b^\star - \tilde b_{\tilde \star})](x)\nonumber\\
&+ \sum_s v_k(s,\tilde \pi_k(s))\sum_{x\in\cS_s^-}
(P_{k} (s,x) - \widetilde P_{k}(s,x))(\tilde b_{\tilde \star}(x) - \tilde b_k (x))\nonumber\\
&- J\sum_s v_k(s,\tilde \pi_k(s))
\sum_{x\in\cS_s^-}
(P_{k} (s,x) - \widetilde P_k(s,x))(g^\star-\tilde { g}_{\tilde \star})\,.\label{eqn:bound1}
\end{align}

We thus get
\begin{align}
\sum_s v_k(s,\tilde \pi_k(s))\bigg(
&(\widetilde P_k- P_k)(\tilde b_k - b^\star)(s)+ g^\star - \tilde g_{\tilde \star}\bigg)\nonumber\\
&\leq
\sum_s v_k(s,\tilde \pi_k(s))\sum_{x\in\cS_s^-}
(P_k (s,x) - \widetilde P_k(s,x))[P_\star^{J}(b^\star- \tilde b_{\tilde \star})](x) + \eta_k \nonumber\\
&+
\sum_s v_k(s,\tilde \pi_k(s))
\bigg[1
-J \sum_{x\in\cS_s^-}
(P_k (s,x) - \widetilde P_k(s,x))
\bigg](g^\star-\tilde { g}_{\tilde \star})\, ,
\label{eqn:bound3}
\end{align}
where $\eta_k := \sum_s
v_k(s,\tilde \pi_k(s))\sum_{x\in\cS_s^-}
(P_{k} (s,x) - \widetilde P_{k}(s,x))(\tilde b_{\tilde\star}(x) - \tilde b_{k}(x))$ is controlled by the error of computing $\tilde b_k$ in episode $k$. In particular, for the considered variant of the algorithm,
\begin{align*}
\eta_k &\le \sum_s v_k(s,\tilde \pi_k(s)) \|p(\cdot|s,\tilde \pi_k(s)) - \tilde p_k(\cdot|s,\tilde \pi_k(s))\|_1 \frac{1}{\sqrt{t_k}} \\
&\le \sqrt{32SB}\sum_{s} \frac{v_k(s,\tilde \pi_k(s))}{N_k(s,\tilde \pi_k(s))^+}\, ,
\end{align*}
where we used $t_k \geq N_k(s,\tilde \pi_k(s))$ for all $s$.

\paragraph{Step 3.} It remains to choose $J$.
To this end, we remark that the mapping induced by $P_\star$ is a contractive mapping, 
namely there exists some $\gamma<1$ such that for any function $f$,
\beqan
\bS(P_\star f) \leq \gamma \bS(f)\,.
\eeqan

Let us choose $J \geq \frac{\log(D)}{\log(1/\gamma)}$, so that with a simple upper bound, it comes
\begin{align*}
\sum_s v_k(s,\tilde \pi_k(s))&\sum_{x\in\cS_s^-}
(P_k(s,x) - \widetilde P_k(s,x))[P_\star^{J}(b^\star- \tilde b_{\tilde \star})](x) \sk
&\leq \sum_s v_k(s,\tilde \pi_k(s)) \|p(\cdot|s,\tilde \pi_k(s)) - \tilde p_k(\cdot|s,\tilde \pi_k(s))\|_1 \frac{\bS(P_\star^{J} (b^\star - \tilde b^\star_{\tilde \star}))}{2} \sk
&\leq \sum_s v_k(s,\tilde \pi_k(s)) \|p(\cdot|s,\tilde \pi_k(s)) - \tilde p_k(\cdot|s,\tilde \pi_k(s))\|_1 De^{-\log(D)} \sk
&\leq \sum_s v_k(s,\tilde \pi_k(s)) \sqrt{\frac{32SB}{N_k(s,\tilde \pi_k(s))^+}}  \; .
\end{align*}

In the sequel, we take $J = \frac{\log(D)}{\log(1/\gamma)}$.
This enables us to control the first two terms in \eqref{eqn:bound3} and it remains to control the term
\begin{align*}
\sum_s v_k(s,\tilde \pi_k(s))
\bigg[
1
-J
\sum_{x\in\cS_s^-}
(P_k(s,x) - \widetilde P_k(s,x))
\bigg]({ g}^\star-\tilde { g}_{\tilde \star})\,.
\end{align*}
In particular we would like to ensure that the term in brackets
is non-negative, since in that case, it is multiplied by a term that is negative. 
To this end,  we note that the term in brackets is lower bounded by
\beqan
1 - J \|p(\cdot|s,\tilde \pi_k(s)) - \tilde p_k(\cdot|s,\tilde \pi_k(s))\|_1 \geq 1 - \frac{\log(D)}{\log(1/\gamma)}  \sqrt{\frac{32SB}{N_k(s,\tilde \pi_k(s))^+}}\,,
\eeqan
and is thus guaranteed to be non-negative since
\beqan
N_k(s,\tilde \pi_k(s))\geq \ell_{s,\tilde \pi_k(s)} \geq 32SB\Big(\frac{\log(D)}{\log(1/\gamma)}\Big)^2\,.
\eeqan

Putting together, we finally have shown that
\begin{align*}
v_k(\widetilde { P}_k - P_k)(\tilde b_k - b^\star)+ v_k(g^\star - \tilde g_k)\mathbf 1&\le v_k(\widetilde { P}_k - P_k)(\tilde b_k - b^\star) + v_k(g^\star - \tilde g_{\tilde \star})\mathbf 1 + \frac{1}{\sqrt{t_k}}v_k \mathbf 1 \\
&\le \bigl(2\sqrt{32SB}  + 1\big)\sum_s \frac{v_k(s,\tilde \pi_k(s))}{\sqrt{N_k(s,\tilde \pi_k(s))^+}} \\
&\le \bigl(2\sqrt{32SB}  + 1\big)\sum_{s,a} \frac{v_k(s,a)}{\sqrt{N_k(s,a)^+}}  \; ,
\end{align*}
which completes the proof.
\ep

\section{Technical Lemmas}
In this section we provide supporting lemmas for the regret analysis. 
The following lemma provides a local version of Pinsker's inequality for two probability distributions, which can be seen as the extension of \citep[Lemma~2]{garivier2016explore} for the case of discrete probability measures.

\begin{lemma}
  \label{lem:refined_Pinsker_vector}
  Let $P$ and $Q$ be two probability distributions on a finite alphabet $\Xcal$. Then,
  \begin{align*}
    \emph{\kl}(P,Q)&\geq   \frac{1}{2}\sum_{x: P(x)\neq Q(x)} \frac{(P(x) - Q(x))^2}{\max (P(x), Q(x))} \, .
  \end{align*}
\end{lemma}
\medskip

\bp
The first and second derivatives of $\kl$ satisfy:
\begin{align*}
\frac{\partial}{\partial P(x)} \kl(P,Q) &= 1+ \log\frac{P(x)}{Q(x)}, \quad \forall x\in \Xcal, \\
\frac{\partial^2}{\partial P(x)\partial P(y)} \kl(P,Q)&= \frac{\bI \{x=y\}}{P(x)}, \quad \forall x,y\in \Xcal.
\end{align*}

By Taylor's Theorem, there exists a probability vector $\Xi$, where $\Xi=tP+(1-t)Q$ for some $t\in (0,1)$, so that
\begin{align*}
\kl(P,Q) &= \kl(Q,Q) + \sum_{x}(P(x)-Q(x))\frac{\partial}{\partial P} \kl(Q,Q) \\
&+ \frac{1}{2}\sum_{x,y} (P(x)-Q(x))(P(y)-Q(y)) \frac{\partial^2}{\partial P(x)\partial P(y)} \kl(\Xi,Q) \\
&= \sum_{x} (P(x) - Q(x)) + \sum_{x} \frac{(P(x)-Q(x))^2}{2\Xi(x)} \\
&\geq  \sum_{x: P(x)\neq Q(x)} \frac{(P(x)-Q(x))^2}{2\max(P(x), Q(x))} \; ,
\end{align*}
thus concluding the proof.
\ep
\medskip

\begin{lemma}[{\citep[Lemma~19]{jaksch2010near}}]
\label{lem:seq_sqrt}
Consider the sequence $(z_k)_{1\leq k\leq n}$ with $0\leq z_k \leq Z_{k-1}:=\max\bigl\{1,\sum_{i=1}^{k-1} z_i\big\}$ for $k\geq  1$ and $Z_0\geq  1$. Then,
\begin{align*}
\sum_{k=1}^n \frac{z_k}{\sqrt{Z_{k-1}}} \leq (\sqrt{2}+1)\sqrt{Z_n}\; .
\end{align*}
\end{lemma}
\medskip

\begin{lemma}
\label{lem:seq_harmonic}
Consider a sequence $(z_k)_{1\leq k\leq n}$ with $0\leq z_k \leq Z_{k-1}:=\max\bigl\{1,\sum_{i=1}^{k-1} z_i\big\}$ for $k\geq  1$ and $Z_0= z_1$. Then,
\begin{align*}
\sum_{k=1}^n \frac{z_k}{Z_{k-1}} \leq 2\log(Z_n) + 1\; .
\end{align*}
\end{lemma}

\bp
We prove the lemma by induction over $n$. For $n=1$, we have $z_1/Z_0 = 1$. Since $Z_1=\max\{1,z_1\}$, it holds that $z_1/Z_0 \leq 2\log(Z_1) + 1$.

Now consider $n>1$. By the induction hypothesis, it holds that $\sum_{k=1}^{n-1} z_k/Z_{k-1} \leq 2\log(Z_{n-1}) +1$. Now it follows from the facts $z_n=Z_{n}-Z_{n-1}$ and $Z_{n-1}\leq Z_n \leq 2Z_{n-1}$ for $n\geq  2$, that
\begin{align*}
\sum_{k=1}^n \frac{z_k}{Z_{k-1}} &\leq 2\log(Z_{n-1}) + \frac{z_n}{Z_{n-1}} + 1 \\
&\leq 2\log(Z_{n-1}) + 2\frac{Z_n - Z_{n-1}}{Z_{n}} + 1 \\
&= 2\log(Z_{n-1}) + 2\Bigl(1-\frac{1}{Z_{n}/Z_{n-1}}\Big) + 1\leq 2\log(Z_n) + 1\; ,
\end{align*}
where the last inequality follows from $\log (x) \geq  1-\frac{1}{x}$ valid for all $x\geq  1$ (see, e.g., \citep{topsoe2006some}). This concludes the proof.
\ep

\begin{lemma}
\label{lem:sum_square_prb}
Let $\alpha_i,\ldots, \alpha_d$ be non-negative numbers and $T\geq  1$, and denote by $V$  the optimal value of the following problem:
\begin{align*}
\max_{x} &\;\; \sum_{i=1}^d \sqrt{\alpha_i x_i} \\
\mathrm{s.t.}&\;\; \sum_{i=1}^d x_i = T \, .
\end{align*}
Then, $V=\sqrt{T\textstyle \sum_{i=1}^d \alpha_i}$.
\end{lemma}
\medskip

\bp
Introduce the Lagrangian
$$
L(x,\lambda) = \sum_{i=1}^d \sqrt{\alpha_i x_i} + \lambda\Bigl(T - \sum_{i=1}^d x_i \Big)\; .
$$
Writing KKT conditions, we observe that the optimal point $x^\star_i, i=1,\ldots,d$ satisfies
$$
\frac{\alpha_i}{2\sqrt{x^\star_i}} - \lambda = 0 ,\;  \forall i, \quad \hbox{ and }\quad  \sum_{i=1}^d x^\star_i - T=0\, .
$$
Hence, we obtain $x_i^\star = \alpha_i/(4\lambda^2)$.  
Plugging this into the equality constraint, it follows that $\lambda = \sqrt{\frac{1}{4T}\textstyle \sum_{j=1}^d \alpha_j}$, thus giving $x^\star_i = \alpha_iT/\sum_{j=1}^d \alpha_j\; $. 
Therefore,
$$
V = \sum_{i=1}^d \sqrt{\alpha_i x^\star_i} = \sum_{i=1}^d \frac{\alpha_i}{\sum_{j=1}^d \alpha_j}\sqrt{T\textstyle \sum_{j=1}^d \alpha_j} = \sqrt{T\textstyle \sum_{j=1}^d \alpha_j},
$$
which completes the proof.
\ep

\section{Background Material on Undiscounted MDPs}\label{app:background}
In this section, we provide the proof of a number of standard results for the sake of self-containedness, and as we believe it helps get intuition on learning in MDPs.
\subsection{Proof of Lemma~\ref{lem:biasgain}}
We provide below a short proof of this standard result for the sake of self-containedness.
\paragraph{The fundamental matrix.}
We first prove the relation involving the fundamental matrix. We note that by direct application of the relation
$	\overline{P}_\pi P_\pi =  P_\pi \overline{P}_\pi =
\overline{P}_\pi \overline{P}_\pi = \overline{P}_\pi$,
it comes
\beqan
(I-P_\pi + \overline{P}_\pi) b_\pi &=& \sum_{t=1}^\infty (I-P_\pi)(P_\pi^{t-1} - \overline{P}_\pi)\mu_\pi
+ \underbrace{\overline{P}_\pi(P_\pi^{t-1} - \overline{P}_\pi)}_{0}\mu_\pi\\
&=&  \sum_{t=1}^\infty (I-P_\pi)P_\pi^{t-1}\mu_\pi - \underbrace{(I-P_\pi)\overline{P}_\pi}_{0}\mu_\pi
= \sum_{t=1}^\infty \big(P_\pi^{t-1}-P_\pi^{t}\big)\mu_\pi\,.
\eeqan
Thus, it remains to show that the latter sum equals $I-\overline{P}_\pi$.
When $P_\pi$ is aperiodic, then the limit $\lim_t P^t_\pi$ exists and is equal to $\overline{P}_\pi$. Thus, we easily get
\beqan
\sum_{t=1}^\infty P_\pi^{t-1}-P_\pi^{t} = \lim_{T\to\infty}(I-P_\pi^T) = I - \lim_{T\to\infty} P_\pi^T = I-\overline{P}_\pi\,.
\eeqan
The general case is more intricate, and we refer to \citep{puterman2014markov}.	

\paragraph{Bellman equation.}	
Now in order to obtain the Bellman equation,  we simply note that	
\beqan
P_\pi b_\pi &=& \sum_{t=1}^\infty (P_\pi^t- \overline{P}_\pi) \mu_\pi =
\sum_{t=2}^\infty (P_\pi^{t-1}- \overline{P}_\pi) \mu_\pi\\
&=& b_\pi - (I-\overline{P}_\pi) \mu_\pi = b_\pi - \mu_\pi + g_\pi\,.
\eeqan	

\subsection{Value Iteration and Stopping Criterion}
\begin{definition}[Value iteration] The value iteration procedure defines a sequence of functions $(u_n)_{n\in\Nat}$ and policies $(\pi_n)_ {n\in\Nat}$ according to the following equations
	\beqan
	\forall n\in \Nat
	\begin{cases}
		u_{n+1}(s)  = \max_{a\in\cA} \mu(s,a) + (P_au_n)(s)\,,
		\quad \text{where } u_0 = 0\\
		\pi_{n+1}(s)  = \cU\Big(\Argmax_{a\in\cA} \mu(s,a) + (P_au_n)(s)\Big)\,,		
	\end{cases}
	\eeqan
	where $\cU(\cB)$  denotes the uniform distribution over a set $\cB$.
\end{definition}

The following result is useful in order to better understand the effect of the classical stopping criterion used for the value iteration procedure.
\begin{lemma}[Value and gain]\label{cor:valuegain}
	Let us assume  that $n$ is such that $\bS(u_{n+1}-u_n) \leq \epsilon$.
	Then it holds that
	\beqan
	g_\star- g_{\pi_{n+1}} \leq \epsilon,
	\quad | u_{n+1} -u_n - g_\star | \leq \epsilon,	
	\quad\text{and}\quad | u_{n+1} -u_n - g_{\pi_n+1} | \leq \epsilon\,.
	\eeqan
\end{lemma}

\bp
We first show that 	the average gain satisfies $ P_\pi g_\pi = g_\pi$ and
\beqan
\forall n\in\Nat,\quad	 \overline{P}_{\pi_{n+1}}[u_{n+1}-u_n] \leq 	g_{\pi_{n+1}}  \leq
g_\star \leq  \overline{P}_{\star}[u_{n+1}-u_n]\,.
\eeqan
Indeed, 	we note that  since
$ \overline{P}_\star P_\star =  \overline{P}_\star$, then for any function $f$, $g_\star = \overline{P}_\star[ \mu_\star + P_{\star} f-f]$. Applying to the function $u_n$, it comes
\beqan
g_\star &=& \overline{P}_\star[ \mu_\star + P_{\star} u_n-u_n]\\
&\leq&  \overline{P}_\star[ \mu_{\pi_{n+1}} + P_{\pi_{n+1}} u_n-u_n]\\
&=& \overline{P}_\star(u_{n+1} - u_n)\,,
\eeqan
where in the second line, we used the maximal property of $\pi_{n+1}$.
On the other hand, we use the equality
\beqan
g_{\pi_{n+1}} = \overline{P}_{\pi_{n+1}}[ \mu_{\pi_{n+1}} + P_{\pi_{n+1}} u_n- u_n] = \overline{P}_{\pi_{n+1}}(u_{n+1}-u_n)\,,
\eeqan
together with the fact that by optimality of $\star$, $g_\star \geq g_{\pi_{n+1}}$.

Thus, all in all it holds on the one hand
\beqan
g_\star- g_{\pi_{n+1}} & \leq& \overline{P}_{\star}[u_{n+1}-u_n] - \overline{P}_{\pi_{n+1}}[u_{n+1}-u_n] \\
&\leq& \max_{s\in\cS}
(u_{n+1}-u_n)(s) - \min_{s\in\cS} [u_{n+1}-u_n] = \bS(u_{n+1}-u_n)\,.
\eeqan
On the  other hand, using similar steps,
\beqan
0 &\leq& \overline{P}_{\star}[u_{n+1}-u_n] - g_\star \leq \max_{s\in\cS}
[u_{n+1}-u_n] - g_\star\\
&\leq& \max_{s\in\cS}
[u_{n+1}-u_n] -  \overline{P}_{\pi_{n+1}}[u_{n+1}-u_n] \leq  \bS(u_{n+1}-u_n)\,.
\eeqan Thus, for all $s\in\cS$, $(u_{n+1}-u_n)(s) - g_\star \leq \epsilon$.
Likewise, we get the reverse inequality $0 \leq g_\star - \min_{s\in\cS}(u_{n+1}-u_n)(s)  \leq \bS(u_{n+1}-u_n) \leq \epsilon$.
The last bound is immediate from
the relation
$	g_{\pi_{n+1}} =\overline{P}_{\pi_{n+1}}(u_{n+1}-u_n)$.	
\ep

\subsection{Pseudo-Regret}
%
%

The following result relates the effective regret to the pseudo-regret
\begin{lemma}[Effective regret to pseudo-regret reduction] Let $\pi$ be any stationary policy. Then it comes for all $T$,
	\beqan\EE[\kR_{\pi,T}(s_1)]
	&=&\big([P_\pi^{T-1}-I]b_\star\big)(s_1)
	 +\sum_{s,a}\Esp[N_T(s,a)] \phi(s,a)\\
	 &\leq& D + \sum_{s,a}\Esp[N_T(s,a)] \phi(s,a)\,.
	\eeqan
\end{lemma}
\bp
	Since $g_\star$ is a constant function, it first comes
	\beqan
	\EE[\kR_{\pi,T}]
	&=& \sum_{t=1}^{T}\Big(g_\star - P_{ \pi}^{t-1} \mu_{ \pi}\Big)=
	 \sum_{t=1}^{T}P_{ \pi}^{t-1} \Big(g_\star - \mu_{ \pi}\Big)\,.
	\eeqan
Then, we note that by construction, it holds that
$g_\star - \mu_\star = (P_\star-I) b_\star$.
Introducing the sub-optimality gap $\phi_{\pi} (s) :=\phi(s,\pi(s))= \mu_\star(s) + (P_\star b_\star)(s) - \mu_\pi(s)- (P_\pi b_\star)(s)$, it then comes
	\beqan
	g_\star - \mu_{ \pi} &=& \phi_{ \pi}
	+ g_\star-\mu_\star -P_\star b_\star + P_{\tilde \pi} b_\star= (P_{ \pi} -I)b_\star  + \phi_{ \pi}\,.
	\eeqan
Thus far, we have we obtained that
\beqan
\EE[\kR_{ \pi,T}]
&=& \sum_{t=1}^{T}P_{ \pi}^{t-1}\phi_{ \pi}
+ \sum_{t=1}^{T}P_{ \pi}^{t-1}(P_{ \pi} -I)b_\star
= \sum_{t=1}^{T}P_{ \pi}^{t-1}\phi_{ \pi} + (P_{ \pi}^{T-1}- I)b_\star\,.
\eeqan
In order to conclude, we  note that
	\beqan
(	\sum_{t=1}^T P_{ \pi_{k}}^{t-1} \phi_{ \pi_k})(s_1) &=& \sum_{t=1}^T \Esp_{s_{t-1}}[\phi_{ \pi_k}(s_{t-1})]\\
	&=& \sum_{s,a} \phi(s,a)\sum_{t=1}^T\Esp_{s_{t-1}}[\ind\{s_{t-1}=s, \pi_k(s)=a\}]
	= \sum_{s,a} \phi(s,a) \Esp[N_T(s,a)]\,.
	\eeqan
For the inequality, we use the simple bound
$[P_\pi^{T-1}-I]b_\star \leq \|P_\pi^{T-1}-I\|_1\frac{1}{2}\bS(b_\star) \leq D\,.$ Putting these together concludes the proof.
\ep

\end{document}